\documentclass[journal]{IEEEtran}
\usepackage{graphicx}
\usepackage{amsmath, amssymb}

\usepackage{caption, subcaption}
\usepackage{multirow}
\usepackage{soul,color}
\usepackage{cite}
\usepackage{hyperref}

\usepackage{tikz}
\usetikzlibrary{shapes, arrows.meta}
\usetikzlibrary{shapes.geometric, arrows, positioning, calc}
\usepackage{pgfplots}
\pgfplotsset{compat=newest}
\begin{document}
\title{DiffFormer: a Differential Spatial-Spectral Transformer for Hyperspectral Image Classification}
\author{Muhammad Ahmad, Manuel Mazzara, Salvatore Distefano, Adil Mehmood Khan, Silvia Liberata Ullo
\thanks{M. Ahmad and S. Distefano are with Dipartimento di Matematica e Informatica-MIFT, University of Messina, 98121 Messina, Italy. (e-mail: mahmad00@gmail.com; sdistefano@unime.it}
\thanks{M. Mazzara is with the Institute of Software Development and Engineering, Innopolis University, Innopolis, 420500, Russia. (e-mail: m.mazzara@innopolis.ru)}
\thanks{A.M. Khan is with the School of Computer Science, University of Hull, Hull HU6 7RX, UK.}
\thanks{S.L. Ullo is with the Engineering Department, University of Sannio, 82100 Benevento, Italy. (e-mail: ullo@unisannio.it)}
}
\markboth{Journal of \LaTeX\ Class Files}
{Ahmad \MakeLowercase{\textit{et al.}}}
\maketitle
\begin{abstract}
Hyperspectral image classification (HSIC) has gained significant attention because of its potential in analyzing high-dimensional data with rich spectral and spatial information. In this work, we propose the Differential Spatial-Spectral Transformer (\textit{DiffFormer}), a novel framework designed to address the inherent challenges of HSIC, such as spectral redundancy and spatial discontinuity. The \textit{DiffFormer} leverages a Differential Multi-Head Self-Attention (DMHSA) mechanism, which enhances local feature discrimination by introducing differential attention to accentuate subtle variations across neighboring spectral-spatial patches. The architecture integrates Spectral-Spatial Tokenization through three-dimensional (3D) convolution-based patch embeddings, positional encoding, and a stack of transformer layers equipped with the SWiGLU activation function for efficient feature extraction (SwiGLU is a variant of the Gated Linear Unit (GLU) activation function). A token-based classification head further ensures robust representation learning, enabling precise labeling of hyperspectral pixels. Extensive experiments on benchmark hyperspectral datasets demonstrate the superiority of \textit{DiffFormer} in terms of classification accuracy, computational efficiency, and generalizability,  compared to existing state-of-the-art (SOTA) methods. In addition, this work provides a detailed analysis of computational complexity, showcasing the scalability of the model for large-scale remote sensing applications. The source code will be made available at \url{https://github.com/mahmad000/DiffFormer} after the first round of revision. 
\end{abstract}
\IEEEpeerreviewmaketitle
\section{Introduction}

\IEEEPARstart{H}{yperspectral Imaging (HSI)} has emerged as a revolutionary technology with diverse applications, ranging from precision agriculture \cite{patro2021review}, object classification \cite{zhai2021hyperspectral}, environmental monitoring \cite{rajabi2024hyperspectral, khan2018modern}, to urban mapping for mineral exploration \cite{peyghambari2021hyperspectral}, just to mention a few. By capturing detailed spectral information for each pixel, HSI enables fine-grained material classification. However, its practical deployment is hindered by challenges such as the high dimensionality, spectral variability, intricate spatial patterns, and the Hughes phenomenon \cite{imani2020overview, bioucas2013hyperspectral, pandey2020future, 10102432, ahmad2021hyperspectral}. Overcoming these obstacles necessitates classification frameworks that are not only accurate but also computationally efficient and capable of effectively leveraging joint spatial-spectral information.

Recent advancements in Transformer-based models have demonstrated remarkable success in HSI classification (HSIC) by utilizing self-attention mechanisms to process image patches effectively \cite{10472541, 10704580, 10571998, 10443948, 10476599, 10379176}. For instance, Huang et al. \cite{10677405} introduce a spectral-spatial vision foundation model-based transformer (SS-VFMT), which augments pre-trained vision foundation models (VFMs) with dedicated spectral and spatial enhancement modules. They further propose a patch relationship distillation strategy (SS-VFMT-D) to optimize pre-trained knowledge utilization and present the spectral-spatial vision-language transformer (SS-VLFMT) for generalized zero-shot classification, enabling the identification of previously unseen classes. While these approaches achieve impressive performance, they face challenges such as high computational complexity and heavy reliance on pre-trained models, which may limit their adaptability to domain-specific applications.

Shu et al. \cite{SHU2024107351} proposed a dual attention transformer network (DATN) for HSIC that integrates a spatial-spectral hybrid transformer (SSHT) module to capture global spatial-spectral dependencies and a spectral local-conv block (SLCB) module to extract local spectral features effectively. While DATN enhances feature representation by combining global and local information, its reliance on multi-head self-attention mechanisms may still lead to computational overhead. Zhong et al. \cite{9565208} proposed a spectral-spatial transformer network (SSTN) that integrates spatial attention and spectral association modules to address convolutional limitations while a factorized architecture search (FAS) framework enables efficient architecture optimization. Despite achieving competitive accuracy with reduced computational costs, relying on a specialized architecture search framework may limit flexibility for broader applications.

Yang et al. \cite{10144788} proposed the Quaternion Transformer Network (QTN), which addresses the limitations of traditional transformers in HSIC by leveraging Quaternion algebra for efficient 3D structure processing and spectral-spatial representation. However, the reliance on hypercomplex computations may increase implementation complexity and resource requirements. Zhang et al. \cite{10189879} proposed a LiT network that integrates lightweight self-attention modules and convolutional tokenization to balance local feature extraction and global dependency capture for HSIC. Despite improved efficiency and reduced overfitting, its reliance on controlled sampling strategies may limit adaptability to diverse datasets. Yang et al. \cite{9766028} proposed a HiT classification network that combines spectral-adaptive 3D convolution and Conv-Permutator modules to enhance spatial-spectral representation in HSIs, addressing Convolutional Neural Networks (CNNs)' limitations in mining spectral sequences. However, the added complexity of these modules may increase computational overhead. Yu et al. \cite{9807344} proposed the MSTNet combining a self-attentive encoder and multilevel features decoding within an efficient transformer-based framework for HSIC. However, reliance on sequence-based processing and positional embeddings may underexploit the inherent 3D structure of HSIs. Zhang et al. \cite{10064266} proposed the MATNet which integrates multi-attention mechanisms and transformers for HSIC, improving boundary pixel classification with spatial-channel attention and a novel Lpoly loss function. However, reliance on semantic-level tokenization may limit fine-grained feature preservation. 

In  \cite{ye2024} novel transformers are introduced addressing noise cancellation able to leverage differential attention. Yet, unfortunately, they exhibit limitations when applied to HSIC. First, the focus on subtractive attention for sparse patterns may overlook the spectral redundancy and spatial discontinuity inherent in HSIC, leading to suboptimal feature discrimination. Additionally, its design is oriented toward text-based tasks, lacking mechanisms like spectral-spatial tokenization and tailored architectures for high-dimensional remote sensing data, which are integral to Spatial-Spectral Transformers (SSTs). Moreover, SSTs face other limitations; for instance, training large SSTs is computationally intensive, particularly due to the quadratic complexity of the self-attention mechanism with respect to sequence length, which hampers scalability \cite{10379170, 10685113, 10419133}. Unlike CNNs with inherent translation invariance from shared weight convolutional filters, SSTs often struggle to robustly capture spatial relationships invariant to small input translations \cite{HUANG2024109897, 10604879, SUN2024102163}. Additionally, reliance on fixed-size patch tokenization can hinder the capture of fine-grained spatial-spectral details \cite{10399798, 10418237, 10400415}. SSTs also require large labeled datasets for optimal performance; on smaller datasets, they risk overfitting, reducing their applicability in data-limited scenarios \cite{10767233}. 

Based on the above analysis of the state of the art (SOTA),to address the aforementioned limitations, this study introduces a novel framework with the following specific contributions:

\begin{enumerate}
    \item \textbf{Differential Attention Mechanism for Localised Feature Discrimination}: a Differential Multi-Head Self-Attention (DMHSA) mechanism that enhances hyperspectral feature representation by computing differential attention scores. These scores capture subtle variations in spectral-spatial regions, enabling the model to highlight localised differences and mitigate the effects of spectral redundancy and spatial noise. This mechanism significantly improves classification accuracy by better modeling inter-pixel relationships.

    \item \textbf{Integration of SWiGLU Activation in Spectral-Spatial Transformers}: The \textit{DiffFormer} incorporates the SWiGLU activation function \cite{articleSWISH} (a variant of the Gated Linear Unit (GLU) activation function) within the feed-forward layers of the transformer blocks. This activation improves non-linear feature transformation, enhancing the network's capacity to capture complex spectral-spatial dependencies while maintaining computational efficiency.

    \item \textbf{Class Token for Unified Spectral-Spatial Representation}: a class token-based aggregation mechanism is utilized, where a dedicated learnable token summarises global spectral-spatial interactions. This approach ensures that the transformer effectively consolidates information from hyperspectral bands, yielding robust feature embeddings for classification tasks. For the above reason, a sinusoidal positional encoding is deployed to hyperspectral data, ensuring precise modeling of spatial and spectral continuity. This tailored encoding is seamlessly integrated with spectral-spatial tokenization, maintaining alignment across bands and patches, a critical factor for HSIC performance.

    \item \textbf{Efficient Patch-Based Spectral-Spatial Tokenization}: The \textit{DiffFormer} employs a 3D convolutional patch embedding strategy, enabling simultaneous extraction of spectral and spatial features in a compact form. This approach reduces the model's input dimensionality while preserving essential spectral-spatial information, enhancing computational efficiency.
\end{enumerate}

This work bridges the gaps identified in SOTA and addresses critical challenges in spectral redundancy, spatial continuity, and computational scalability, by setting a new paradigm for HSIC models tailored for real-time applications.

\section{Proposed Methodology}

\begin{figure*}[!hbt]
    \centering
    \includegraphics[width=0.99\linewidth]{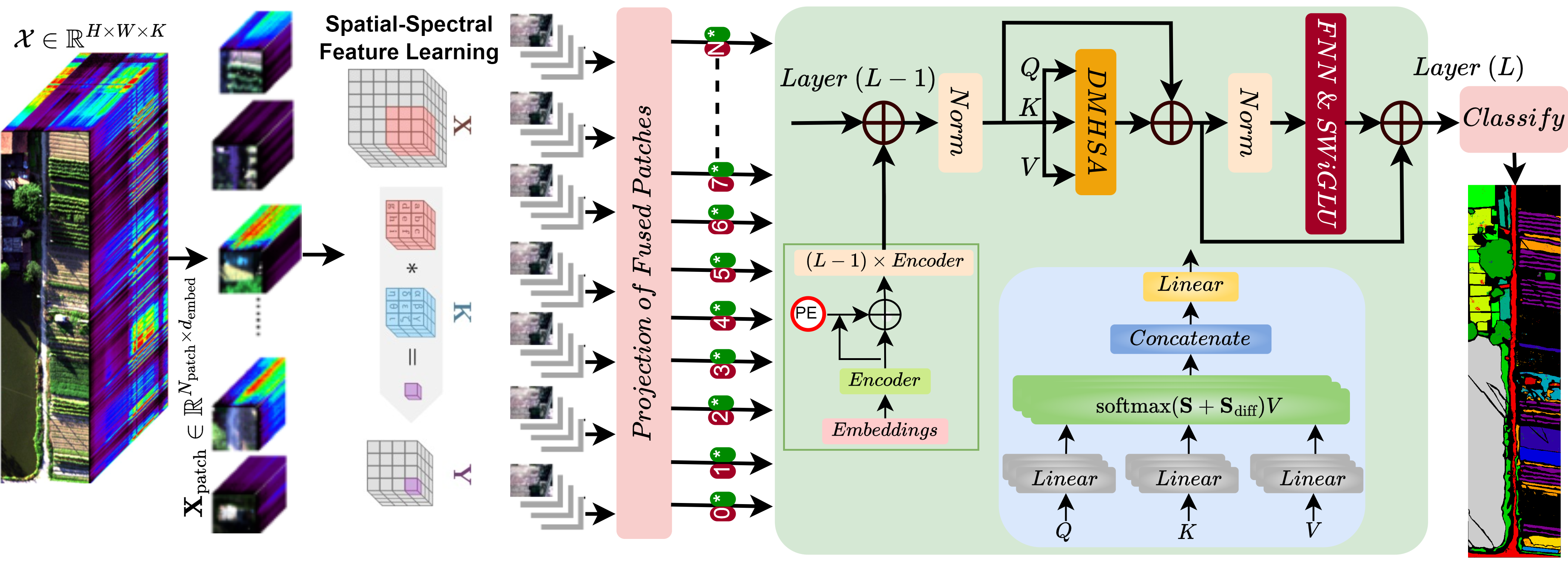}
    \caption{Schematic representation of the DiffFormer pipeline for HSIC. The pipeline starts with hyperspectral data preprocessing, where fused patches are generated and spatial-spectral features are extracted. Differential Attention is employed within the encoder to refine the attention mechanism by integrating positional embeddings (PE) for enhanced spectral-spatial relationships. The hierarchical encoding layers aggregate the learned features across $L-1$ layers, enabling multi-scale representation learning. The model's effectiveness is evaluated on HSIC tasks, demonstrating its ability to accurately delineate class boundaries in hyperspectral datasets.}
    \label{Fig1}
\end{figure*}

This section presents the proposed Differential Spatial-Spectral Transformer (\textit{DiffFormer}) model for HSIC. The model leverages differential attention mechanisms and spatial-spectral tokenization to effectively encode both spatial and spectral dependencies. The architectural components and the proposed innovations are presentec in Figure \ref{Fig1}. The \textit{DiffFormer} is designed to capture complex spatial and spectral relationships in HSIs through an integration of convolutional and transformer-based methodologies. The architecture includes the following core components: spatial-spectral patch embedding, positional encoding, a Differential Multi-Head Self-Attention (DMHSA) module, and a classification head.

The input hyperspectral cube, of dimensions $(W, H, K)$, where $W$ and $H$ are the spatial dimensions and $K$ represents the number of spectral bands, undergoes spatial-spectral patch embedding. A 3D convolutional layer extracts feature representations using patches of size $P \times P \times K$. The resulting feature tensor is reshaped into a sequence of patch embeddings:

\begin{equation}
    \mathbf{X}_\text{patch} \in \mathbb{R}^{N_\text{patch} \times d_\text{embed}}
\end{equation}
where $N_\text{patch}$ is the number of patches, and $d_\text{embed}$ is the embedding dimension. Additionally, a trainable class token is appended to this sequence.

\subsubsection{\textbf{Positional Encoding}}

To retain spatial context, a sinusoidal positional encoding is added to the patch embeddings:

\begin{equation}
    \mathbf{X}_\text{pos} = \mathbf{X}_\text{patch} + \mathbf{PE}
\end{equation}
where $\mathbf{PE} \in \mathbb{R}^{(N_\text{patch}+1) \times d_\text{embed}}$ encodes spatial relationships. The positional encoding matrix is precomputed using trigonometric functions:

\begin{equation}
    \text{PE}(pos, 2i) = \sin\left(\frac{pos}{10000^{2i/d_\text{embed}}}\right)
\end{equation}

\begin{equation}
    \text{PE}(pos, 2i+1) = \cos\left(\frac{pos}{10000^{2i/d_\text{embed}}}\right)
\end{equation}

\subsubsection{\textbf{Differential Multi-Head Self-Attention (DMHSA)}}

The core innovation of the \textit{DiffFormer} lies in the DMHSA module. DMHSA extends traditional multi-head attention by introducing a differential operation on attention scores to capture relative changes between neighboring tokens:

\begin{equation}
    \mathbf{S} = \frac{\mathbf{Q}\mathbf{K}^\top}{\sqrt{d_\text{head}}}
\end{equation}

\begin{equation}
    \mathbf{S}_\text{diff} = \mathbf{S}[:, 1:] - \mathbf{S}[:, :-1]
\end{equation}
where $\mathbf{Q}, \mathbf{K}, \mathbf{V}$ are the query, key, and value matrices, and $d_\text{head}$ is the head dimension. The \(\mathbf{S}_\text{diff}\) introduces several benefits: First, it captures relative changes in attention scores, emphasizing the dynamics of transitions between consecutive tokens rather than focusing solely on absolute similarities. This is particularly useful for identifying shifts in importance across tokens. Additionally, \(\mathbf{S}_\text{diff}\) enhances sensitivity to local variations by highlighting regions where attention shifts significantly, which is crucial for tasks requiring fine-grained context understanding, such as HSIC. By suppressing uniform or redundant attention scores, the operation produces sparse and interpretable attention maps, reducing computational overhead while improving explainability. Furthermore, \(\mathbf{S}_\text{diff}\) increases robustness to noise, as it is less affected by small perturbations in query-key interactions, making it ideal for noisy data scenarios. It also naturally encodes sequential dependencies by focusing on transitions between tokens, which is valuable for sequence modeling and applications with strong temporal or spatial correlations. Moreover, the differential attention mechanism facilitates multi-scale representations when combined with hierarchical or multi-head attention, capturing both absolute and relative importance at different granularities. Finally, \(\mathbf{S}_\text{diff}\) can be employed as a regularization to ensure smooth transitions in attention scores, thereby mitigating abrupt changes caused by overfitting, which enhances the overall generalization of the model.

The resulting attention weights are used to compute the weighted sum of values:

\begin{equation}
    \mathbf{Z} = \mathbf{A}\mathbf{V}
\end{equation}

\subsubsection{\textbf{Transformer Encoder Block}}

The transformer encoder block, implemented as a SST layer, incorporates the DMHSA module along with feedforward layers enhanced by SWiGLU activation:

\begin{equation}
    \text{SWiGLU}(x, g) = x \odot \sigma(g) + x
\end{equation}
where $\sigma(\cdot)$ denotes the sigmoid function. Layer normalization and residual connections ensure stable training. The output from the final SST layer is passed through a dense layer to extract the class token representation. This representation is fed into fully connected layers with L2 regularization to generate the classification logits:

\begin{equation}
    \mathbf{y} = \text{softmax}(\mathbf{W}\mathbf{z}_\text{cls} + \mathbf{b})
\end{equation}
where $\mathbf{z}_\text{cls}$ is the class token embedding, and $\mathbf{W}, \mathbf{b}$ are trainable parameters.

The model is trained using categorical cross-entropy loss and optimized with the Adam optimizer. The following metrics are computed for evaluation:  Overall Accuracy (OA): The percentage of correctly classified samples. Average Accuracy (AA): The mean accuracy across all classes. Kappa Coefficient ($\kappa$): A measure of agreement between predictions and ground truth. Performance is further analyzed using classification reports and confusion matrices.

\subsubsection{\textbf{Computational Complexity and Implementation}}

The proposed \textit{DiffFormer} achieves a balance between computational efficiency and accuracy. The attention mechanism introduces a computational cost of $O(N_\text{patch}^2 \cdot d_\text{head})$, while the integration of differential attention and SWiGLU enhances feature expressiveness with minimal overhead. The implementation is conducted in TensorFlow, leveraging GPU acceleration for efficient processing.

\section{Experimental Datasets and Settings}

To validate the efficacy of the \textit{DiffFormer}, we utilize real-world HSI datasets characterized by diverse spatial and spectral features. This section details the datasets, experimental setup, evaluation metrics, and comparative results to demonstrate the robustness and adaptability of the \textit{DiffFormer}.

The \textbf{WHU-Hi-HanChuan (HC)} dataset \cite{zhong2018mini} was collected between 17:57 and 18:46 on 17 June 2016 in Hanchuan, Hubei Province, China. The data was acquired using a Headwall Nano-Hyperspec imaging sensor with a 17-mm focal length, mounted on a Leica Aibot X6 UAV V1 platform. The weather conditions were ideal for data collection, with clear skies, a temperature of approximately 30$^\circ\mathrm{C}$, and relative humidity around 70\%. The study area represents a fringe zone of rural and urban encompassing various types of land cover, including buildings, water bodies, and cultivated fields with seven different crop species: strawberry, cowpea, soybean, sorghum, water spinach, watermelon, and greens. The UAV operated at an altitude of 250 m, capturing hyperspectral imagery with a spatial resolution of 0.109 m and a size of 1217 $\times$ 303 pixels. The dataset comprises 274 spectral bands spanning wavelengths from 400 to 1000 nm. It is worth noting that the imagery was collected in the late afternoon when the solar elevation angle was low, resulting in the presence of significant shadow-covered regions within the dataset. This aspect adds complexity to the analysis and highlights the challenges of real-world HSIC.

The 
(\href{https://www.ehu.eus/ccwintco/index.php/Hyperspectral_Remote_Sensing_Scenes}{\textbf{Salinas (SA)} dataset} is a widely used benchmark in HSI analysis, acquired using the 224-band Airborne Visible/Infrared Imaging Spectrometer (AVIRIS) sensor. Data were collected over Salinas Valley, California, a region known for its agricultural diversity. This dataset features high spatial resolution imagery with pixel sizes of 3.7 meters, providing detailed information suitable for land cover classification and analysis. The captured scene spans an area of 512 lines by 217 samples, covering various land cover types, including diverse vegetable crops, bare soil regions, and vineyard fields. To ensure data quality, bands affected by water absorption, specifically bands 108–112, 154–167, and 224, were excluded from the analysis, leaving 204 usable spectral bands. It is important to note that the dataset was made available as at-sensor radiance data, without atmospheric correction applied, which may add challenges to its processing and interpretation. The ground truth for the Salinas dataset consists of 16 distinct land cover classes, reflecting the agricultural and environmental characteristics of the region. These classes provide a diverse set of spectral and spatial patterns, making this dataset ideal for evaluating the performance of HSIC models.

The 
(\href{https://www.ehu.eus/ccwintco/index.php/Hyperspectral_Remote_Sensing_Scenes}{\textbf{Pavia University (PU)} dataset} 
was captured by the Reflective Optics System Imaging Spectrometer (ROSIS) sensor during a flight campaign over the University of Pavia, located in northern Italy. This dataset has been widely used in HSIC studies due to its high spatial and spectral resolution. The image consists of 610 $\times$ 610 pixels with a geometric resolution of 1.3 meters per pixel, offering fine-grained spatial details suitable for analyzing urban environments. It contains 103 spectral bands across the visible and near-infrared portions of the spectrum. However, certain spectral bands may need pre-processing to remove unwanted noise or artifacts, and some regions in the image contain no meaningful information, appearing as black strips in the dataset. These areas should be excluded from the analysis to focus on relevant data. The ground truth associated with the PU dataset categorizes the image into 9 distinct land cover classes. These include features typical of urban and peri-urban landscapes, providing a diverse and challenging classification scenario. The dataset is frequently utilized to benchmark algorithms for urban mapping and land cover classification, making it an essential resource in HSI research. Its high-quality spectral and spatial data, combined with the complexities introduced by the non-informative regions, make the PU dataset an excellent testbed for developing and evaluating advanced hyperspectral processing techniques.

The \textbf{University of Houston (UH)} dataset, introduced by the IEEE Geoscience and Remote Sensing Society (GRSS) during the 2013 Data Fusion Contest \cite{debes2014hyperspectral}, is a benchmark hyperspectral dataset widely used in remote sensing research. Acquired using the Compact Airborne Spectrographic Imager (CASI), the dataset captures an urban environment in Houston, Texas, with exceptional spectral and spatial resolution. The HSI features a spatial resolution of 2.5 meters per pixel and dimensions of 340 $\times$ 1,905 pixels, covering a diverse urban and semi-urban area. It includes 144 spectral bands spanning wavelengths from 0.38 to 1.05 $\mu m$, encompassing the visible spectrum and portions of the near-infrared region. This combination of high spatial and spectral detail makes it ideal for distinguishing subtle variations in land-cover types. The dataset is accompanied by ground truth with 15 distinct land-cover classes, including buildings, roads, parking lots, cars, shadows, grass, trees, water bodies, bare soil, synthetic materials, urban vegetation, residential areas, commercial zones, mixed vegetation, and other impervious surfaces. As a rich testbed for advancing HSI analysis, this dataset has become a cornerstone in remote sensing research. Table \ref{Tab1} provides the experimental datasets' key details and characteristics.

\begin{table}[!hbt]
    \centering
    \caption{Summary of the HSI datasets used for experimental evaluation.}
    \resizebox{\columnwidth}{!}{\begin{tabular}{ccccc} \hline 
        --- & \textbf{SA} & \textbf{UH} & \textbf{PU} & \textbf{HC} \\  \hline 
        \textbf{Sensor} & AVIRIS & CASI & ROSIS-03 & Headwall Nano \\
        \textbf{Wavelength} & $350-1050$ & $350-1050$ & $430-860$ & $400-1000$ \\
        \textbf{Resolution} & $3.7m$ & $2.5m$ & $1.3m$ & $0.109m$ \\ 
        \textbf{Spatial} & $512 \times 217$ & $340\times 1905$ & $610 \times 610$ & $1217 \times 303$ \\
        \textbf{Spectral} & 224 & 144 & 103 & 274 \\
        \textbf{Classes} & 16 & 15 & 9 & 16 \\
        \textbf{Source} & Aerial & Aerial & Aerial & Aerial \\ \hline 
    \end{tabular}}
    \label{Tab1}
\end{table}

The experimental settings for evaluating the \textit{DiffFormer} were designed to ensure a robust assessment of its performance across the different datasets. In this setup, the input data is processed using patches with a spatial size of 8, which defines the extent of each input sample for training and testing. To reduce the dimensionality of the spectral data, principal component analysis (PCA) is employed, selecting the top 15 spectral bands that contribute most significantly to the variance, thereby retaining the most relevant information for classification.

The dataset is partitioned into three subsets: 50\% for testing 25\% for validation, and 25\% for training. This split ensures a balanced approach, allowing for both proper model training and robust evaluation. During the training phase, the Adam optimizer from the TensorFlow legacy module is used with a learning rate of 0.001 and a decay rate of $1 \times 10^{-6}$, helping the model converge efficiently while minimizing the risk of overfitting. The training process spans 50 epochs, with a batch size of 56, which is large enough to allow for efficient gradient updates without overwhelming the system's memory capacity.

The model architecture is based on a transformer framework, comprising 4 transformer layers, each equipped with 8 attention heads. The feed-forward network within each transformer layer consists of $4 \times 64$ units, enabling the model to extract rich and complex features from the input data. To mitigate overfitting, dropout is applied with a rate of 0.1, and layer normalization is incorporated with an epsilon value of $1 \times 10^{-3}$ to stabilize the learning process. Additionally, a kernel regularizer with an L2 penalty of 0.01 is applied to the model's weights to further promote generalization and avoid overfitting.

After training and validating the model, its performance is evaluated on the test set using several key metrics: OA, which reflects the proportion of correctly classified samples across all classes; AA, which provides the mean classification accuracy per class; Per-Class Accuracy, which evaluates the performance of the model for each individual class; and the $\kappa$, a measure that accounts for agreement between the predicted and true labels, adjusted for chance. These metrics together provide a comprehensive assessment of the model’s effectiveness in various aspects.

\section{Impact of Patch sizes}

The choice of patch size plays a crucial role in the performance of HSIC models, as it directly influences both the spatial and spectral information that can be captured by the model. Patch size determines the extent of the local context considered during training and testing, which in turn impacts the ability of the model to capture fine-grained spatial features as well as the spectral variations present in the data. Smaller patch sizes may enable the model to focus on finer details but can lead to loss of contextual information, while larger patch sizes allow for more comprehensive spatial information but might reduce the ability to distinguish subtle spectral differences. In this section, we explore the impact of varying patch sizes on classification performance, examining how patch size influences accuracy, model convergence, and the ability to generalize to unseen data. Through a series of experiments, we assess the trade-offs between spatial detail and computational complexity, providing insights into how patch size selection can optimize performance in HSIC tasks.

\begin{table}[!hbt]
    \centering
    \caption{Classification performance of different patch sizes evaluated on four datasets. The highest values for each metric highlighted in bold. The results demonstrate the impact of patch size on classification performance, with larger patch sizes generally yielding better results across most datasets.}
    \resizebox{\columnwidth}{!}{\begin{tabular}{c||ccc||ccc||ccc||ccc} \hline 
        \multirow{2}{*}{\textbf{Patch}} & \multicolumn{3}{c||}{\textbf{HC}} & \multicolumn{3}{c||}{\textbf{UH}} & \multicolumn{3}{c||}{\textbf{SA}} & \multicolumn{3}{c}{\textbf{PU}} \\ \cline{2-13}
        & \textbf{$\kappa$} & \textbf{OA} & \textbf{AA} & \textbf{$\kappa$} & \textbf{OA} & \textbf{AA} & \textbf{$\kappa$} & \textbf{OA} & \textbf{AA} & \textbf{$\kappa$} & \textbf{OA} & \textbf{AA} \\ \hline 
        
        $8 \times 8$ & 95.20 & 95.90 & 91.27 & 97.29 & 97.49 & 96.88 & 98.20 & 98.38 & 99.23 & 96.87 & 97.64 & 96.58 \\ 
        $10 \times 10$ & 96.56 & 97.06 & 93.65 & 96.81 & 97.05 & 96.03 & 98.76 & 98.89 & 99.41 & 97.54 & 98.15 & 97.35 \\ 
        $12 \times 12$ & 96.34 & 96.87 & 93.55 & 97.38 & 97.57 & 97.08 & 99.13 & 99.22 & 99.55 & 96.76 & 97.56 & 96.90 \\ 
        $14 \times 14$ & 97.09 & 97.52 & 95.29 & 97.39 & 97.59 & 96.76 & 99.06 & 99.15 & 99.48 & \textbf{98.00} & \textbf{98.49} & \textbf{97.41} \\ 
        $16 \times 16$ & 96.50 & 97.01 & 92.48 & 96.61 & 96.87 & 96.29 & \textbf{99.53} & \textbf{99.58} & \textbf{99.76} & 97.37 & 98.01 & 97.54 \\ 
        $18 \times 18$ & \textbf{97.26} & \textbf{97.66} & \textbf{95.65} & 80.85 & 82.31 & 77.89 & 99.48 & 99.54 & 99.77 & 97.98 & 98.48 & 97.65 \\ 
        $20 \times 20$ & 96.56 & 97.06 & 93.65 & \textbf{97.91} & \textbf{98.07} & \textbf{97.90} & 99.06 & 99.16 & 99.16 & 97.77 & 98.32 & 97.25 \\ \hline 
    \end{tabular}}
    \label{Tab2}
\end{table}
\begin{figure}[!hbt]
    \begin{tikzpicture}
    \begin{axis}[
        width=0.45\textwidth, 
        height=0.30\textwidth, 
        ylabel={Time (s)},
        grid=both,
        grid style={dashed, gray!30},
        legend style={
            at={(0.5,0.55)},
            anchor=north, 
            font=\small,
            legend columns=4,
            /tikz/every even column/.append style={column sep=0.2cm} 
        },
        xtick={1,2,3,4,5,6,7},
        xticklabels={$8 \times 8$,$10 \times 10$,$12 \times 12$,$14 \times 14$,$16 \times 16$,$18 \times 18$,$20 \times 20$},
        mark options={solid},
        cycle list name=color list,
        xticklabel style={rotate=45, anchor=north east}
    ]
            \addplot[color=blue, thick, mark=square*] coordinates {(1, 5248) (2, 5964) (3, 5235) (4, 5157) (5, 6183) (6, 5370) (7, 5560)};
            \addplot[color=red, thick, mark=triangle*] coordinates {(1, 322) (2, 322) (3, 313) (4, 318) (5, 318) (6, 323) (7, 326)};
            \addplot[color=green, thick, mark=*] coordinates {(1, 1104) (2, 1115) (3, 1110) (4, 1116) (5, 1129) (6, 1100) (7, 1139)};
            \addplot[color=black, thick, mark=o] coordinates {(1, 877) (2, 890) (3, 893) (4, 877) (5, 912) (6, 943) (7, 889)};
            \legend{HC, UH, SA, PU}
            \end{axis}
    \end{tikzpicture}
    \caption{Time comparison for different patch sizes across four datasets. The x-axis represents the patch size, while the y-axis denotes the time taken for processing in seconds.}
    \label{Fig2}
\end{figure}
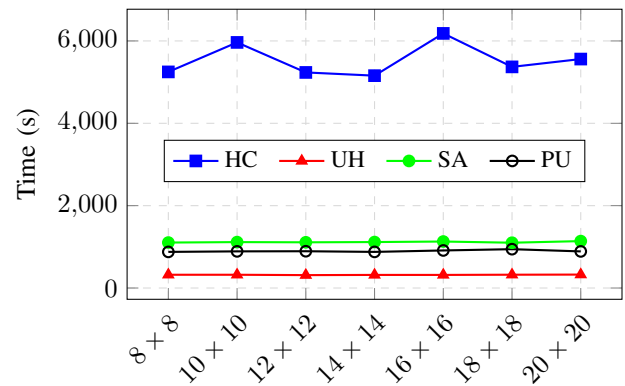

Table \ref{Tab2} and Figure \ref{Fig2} present the classification performance and computational time for different patch sizes across four datasets: HC, UH, SA, and PU. The results are evaluated using three metrics: $\kappa$, OA, and AA. The values in bold represent the highest results within each dataset and metric. For the HC dataset, the best performance across all metrics is observed with the patch size of $18 \times 18$. While the performance metrics for smaller patch sizes such as $8 \times 8$ and $10 \times 10$ are quite competitive, the $18 \times 18$ patch size leads to the highest $\kappa$ (97.91), OA (98.07), and AA (97.90), indicating superior classification performance when larger patches are considered. This trend is consistent across all datasets to some extent, suggesting that larger patch sizes may provide better spatial context for classification.

For the UH dataset, the $14 \times 14$ patch size yields strong results, with $\kappa$, OA, and AA values reaching 97.39, 97.59, and 96.76, respectively. However, the $20 \times 20$ patch size slightly improves Kappa and OA, reaching 97.91 and 98.07, making it the best-performing configuration in this case as well. This suggests that the model benefits from larger patches for this dataset, capturing more contextual information that leads to enhanced performance. The SA dataset also shows that the larger patch sizes, particularly $12 \times 12$ and $14 \times 14$, perform well. The $14 \times 14$ patch size provides a good balance with Kappa, OA, and AA values of 99.06, 99.15, and 99.48, respectively. However, the largest patch size, $18 \times 18$, gives the best performance with a $\kappa$ of 99.53, OA of 99.58, and AA of 99.76, indicating that the model is better able to classify the different land cover types when larger spatial contexts are considered.

In the case of the PU dataset, the performance is also highest for larger patch sizes. The $14 \times 14$ patch provides strong results, but the best overall performance is achieved by the $18 \times 18$ patch size, which shows a significant increase in $\kappa$ (97.26), OA (97.66), and AA (95.65). Notably, the $18 \times 18$ patch size significantly outperforms the smaller patch sizes, indicating that it captures sufficient spatial context for better classification accuracy. Overall, these results suggest that larger patch sizes generally provide better classification performance across most datasets, particularly in terms of $\kappa$ and OA. The $18 \times 18$ and $20 \times 20$ patch sizes tend to perform better in capturing the spatial context of the datasets, though in some cases, such as with the UH dataset, smaller patch sizes still offer competitive performance. The choice of optimal patch size may thus depend on the specific characteristics of the dataset and the trade-off between computational cost and accuracy.

\section{Impact of Training Samples}

In HSIC, the selection and amount of training samples play a crucial role in determining the performance of a model. The availability of representative training data significantly influences the model's ability to generalize well to unseen samples. In this section, we explore the impact of the distribution of training samples on the classification results, focusing on how different sample sizes can affect the accuracy, robustness, and efficiency of the model. Through experimental evaluations, we investigate the trade-offs between sample size and model performance, including the effects of overfitting and underfitting when training with insufficient or excessive samples. The findings aim to provide insights into the optimal sample size required for achieving the best performance while maintaining computational efficiency.

\begin{figure}[!hbt]
    \begin{tikzpicture}
    \begin{axis}[
        width=0.49\textwidth, 
        height=0.30\textwidth, 
        ylabel={OA},
        grid=both,
        grid style={dashed, gray!30},
        legend style={
            at={(0.6,0.20)},
            anchor=north, 
            font=\small,
            legend columns=4,
            /tikz/every even column/.append style={column sep=0.2cm} 
        },
        xtick={1,2,3,4,5,6},
        xticklabels={$7\%$,$14\%$,$21\%$,$28\%$,$35\%$,$42\%$},
        mark options={solid},
        cycle list name=color list]
            \addplot[color=blue, thick, mark=square*] coordinates {(1, 93.7065) (2, 95.2821) (3, 96.0069) (4, 97.1731) (5, 98.1167) (6, 98.1426)};
            \addplot[color=red, thick, mark=triangle*] coordinates {(1, 93.2357) (2, 96.8285) (3, 95.8305) (4, 96.6693) (5, 96.9616) (6, 98.6693)};
            \addplot[color=green, thick, mark=*] coordinates {(1, 97.6784) (2, 98.7991) (3, 98.8238) (4, 99.5319) (5, 99.6319) (6, 99.7472)};
            \addplot[color=black, thick, mark=o] coordinates {(1, 93.7426) (2, 96.2908) (3, 97.8804) (4, 98.44151) (5, 97.7246) (6, 98.8934)};
            \legend{HC, UH, SA, PU}
            \end{axis}
    \end{tikzpicture}
    \caption{Classification performance of different percentage of training samples. The results demonstrate the impact of training samples with $12 \times 12$ patch size.}
    \label{Fig3}
\end{figure}
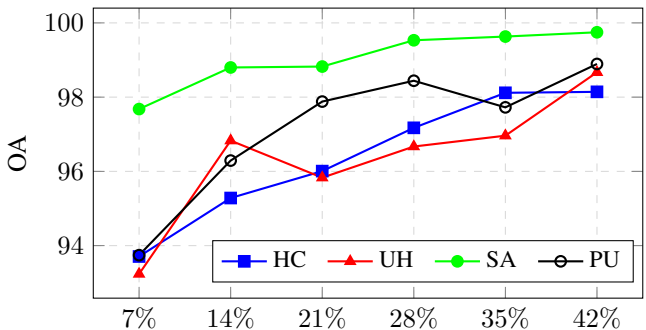

Figure \ref{Fig3} illustrates the classification performance of different percentages of training samples using a patch size $12 \times 12$. The y-axis represents the OA, measured in percentage, while the x-axis denotes the different percentages of the training sample: 7\%, 14\%, 21\%, 28\%, 35\%, and 42\%. The results indicate a clear trend in classification accuracy relative to the percentage of training samples utilized. As the proportion of training samples increases, there is a noticeable enhancement in OA for all four datasets. This trend underscores the importance of sufficient training data in achieving optimal model performance. Among the datasets, the SA dataset consistently outperforms the others across all sample percentages, achieving an OA that exceeds 99\% when using 35\% and 42\% of the training data. This superior performance suggests that the SA dataset may possess more informative features or better representational characteristics that facilitate higher accuracy in classification tasks. The figure also highlights diminishing returns on accuracy gains as the sample size increases beyond a certain threshold. For instance, while moving from 28\% to 35\% shows a significant increase in OA, further increases to 42\% yield only marginal improvements. This phenomenon is critical for practical applications, as it indicates that after reaching an optimal sample size, additional data may not substantially enhance model performance but could lead to increased computational costs and time.

\section{Impact of Transformer Layers}

The number of transformer layers in a model significantly influences its ability to capture complex spatial and spectral dependencies in HSIC. Transformer-based architectures, with their attention mechanisms, allow for the modeling of long-range relationships between pixels, making them particularly effective for high-dimensional data such as HSIs. In this section, we explore the impact of varying the number of transformer layers on the model's performance. We examine how the depth of the model affects its capacity to learn intricate patterns, while also considering potential trade-offs such as increased computational cost and risk of overfitting. Through a series of experiments, we aim to determine the optimal number of transformer layers that balance performance and efficiency, providing insights into the architectural choices that best suit HSIC.

\begin{figure}[!hbt]
    \centering
    \begin{tikzpicture}
        \begin{axis}[
            width=0.48\textwidth, 
            height=0.30\textwidth, 
            ybar,
            grid=both,
            grid style={dashed, gray!30},
            xlabel={Layers},
            ylabel={OA},
            xtick={1,2,3,4,5,6},
            legend style={
                at={(0.5,1.17)},
                anchor=north, 
                font=\small,
                legend columns=4,
                /tikz/every even column/.append style={column sep=0.2cm}},
            bar width=0.15,
        ]
            \addplot coordinates {(1,98.58) (2,98.61) (3,98.26) (4,98.14) (5,97.67) (6,94.84)};
            \addplot coordinates {(1,99.26) (2,97.24) (3,98.29) (4,96.96) (5,98.13) (6,95.71)};
            \addplot coordinates {(1,99.57) (2,99.19) (3,99.49) (4,99.34) (5,99.47) (6,99.28)};
            \addplot coordinates {(1,98.79) (2,98.83) (3,98.82) (4,98.89) (5,98.23) (6,98.30)};
            \legend{HC, UH, SA, PU}
        \end{axis}
    \end{tikzpicture}
    \caption{Impact of transformer layer depth on OA for HSIC. The bar plot shows the OA achieved by the DiffFormer model across six transformer layers for four datasets using a patch size of 12 $\times$ 12. }
    \label{Fig4}
\end{figure}
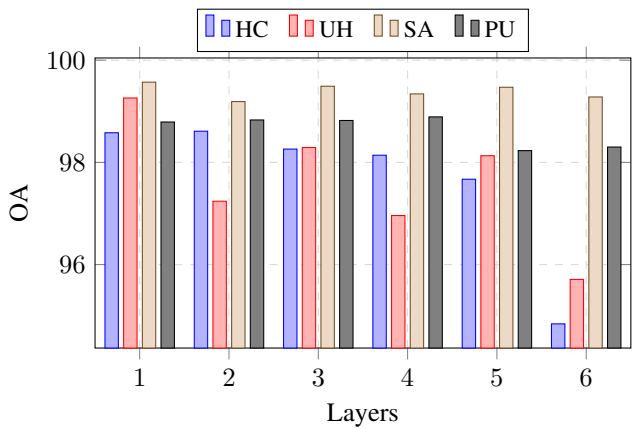

The results presented in Figure \ref{Fig4} offer valuable insights into the performance of the \textit{DiffFormer} across various transformer layer configurations for four distinct hyperspectral datasets. Using a consistent patch size of 12 $\times$ 12, the model's OA was evaluated for layer depths ranging from 1 to 6, revealing complex interactions between model architecture and dataset characteristics. The SA dataset demonstrates remarkable performance consistency, maintaining OA above 99\% across all layer configurations. This suggests that the SA data set's spatial and spectral features are particularly well suited to the \textit{DiffFormer} architecture, allowing for robust classification even with minimal layer depth. In contrast, the HC, UH, and PU datasets exhibit more varied responses to increasing layer depth. For these datasets, peak performance is generally observed with fewer layers (1-3), followed by a gradual decline in accuracy as the layer count increases. This trend indicates that, for most datasets, the model can effectively capture relevant features with a relatively shallow architecture, and additional layers may introduce unnecessary complexity or overfitting.

Interestingly, each dataset presents a unique performance pattern across layers, underscoring the importance of dataset-specific model tuning. For instance, the HC dataset shows a sharp decline in performance at 6 layers, while the UH dataset exhibits more fluctuation across different layer counts. The PU dataset maintains relatively stable performance across all configurations, suggesting a certain robustness to architectural changes for this particular data. These findings highlight the nuanced relationship between model depth and classification accuracy in HSI analysis. They emphasize the need for careful consideration of dataset characteristics when designing and optimizing deep learning models for such tasks. Moreover, the results suggest that in many cases, simpler models with fewer layers may be sufficient or even preferable for achieving high classification accuracy, potentially offering benefits in terms of computational efficiency and reduced risk of overfitting.

\section{Impact of Attentional Heads}

The number of attention heads in a transformer model plays a crucial role in capturing diverse and complex relationships within the input data. Attention heads allow the model to focus on different parts of the input simultaneously, enabling it to learn multiple data representations in parallel. In the context of HSIC, where both spatial and spectral correlations are essential, the choice of attention heads can significantly impact the model's ability to capture fine-grained features. This section delves into the effect of varying the number of attention heads on the model's performance. We investigate how the capacity to attend to multiple aspects of the data simultaneously influences classification accuracy and model efficiency, while also addressing potential challenges such as the trade-off between model complexity and generalization. Through detailed experiments, we aim to identify the optimal configuration of attention heads that maximizes the performance of \textit{DiffFormer}.

\begin{figure}[!hbt]
    \centering
        \begin{tikzpicture}
        \begin{axis}[
        xlabel={Heads},
        ylabel={Time (s)},
        zlabel={OA (\%)},
        grid=both,
        colormap/viridis, 
        view={30}{30},
        scatter/classes={
            HC={mark=*,red}, 
            UH={mark=square*,blue},
            PU={mark=triangle*,green},
            SA={mark=diamond*,purple}
        },
        ytick={1,10,20,30,40},
    ]
    \addplot3[
        only marks,
        scatter, 
        mark size=4pt,
        scatter src=explicit symbolic
    ]
    coordinates {
        (1, 42.76, 98.8079) [HC]
        (2, 42.02, 98.6124) [HC]
        (4, 40.31, 97.6805) [HC]
        (8, 44.62, 98.6150) [HC]

        (1, 1.91, 97.6713) [UH]
        (2, 1.73, 97.8487) [UH]
        (4, 1.89, 98.3588) [UH]
        (8, 2.08, 99.2681) [UH]

        (1, 2.08, 99.2681) [PU]
        (2, 8.07, 98.6597) [PU]
        (4, 7.98, 98.8155) [PU]
        (8, 8.26, 98.8311) [PU]

        (1, 6.85, 99.3472) [SA]
        (2, 5.55, 99.4580) [SA]
        (4, 5.13, 99.4149) [SA]
        (8, 6.40, 99.5750) [SA]
    };
    \legend{HC, UH, PU, SA}
    \addlegendimage{mark=*,red} 
    \addlegendimage{mark=square*,blue} 
    \addlegendimage{mark=triangle*,green} 
    \addlegendimage{mark=diamond*,purple}
    \end{axis}
    \end{tikzpicture}
    \caption{Overall Accuracy and Time for Different Heads}
    \label{Fig5}
\end{figure}
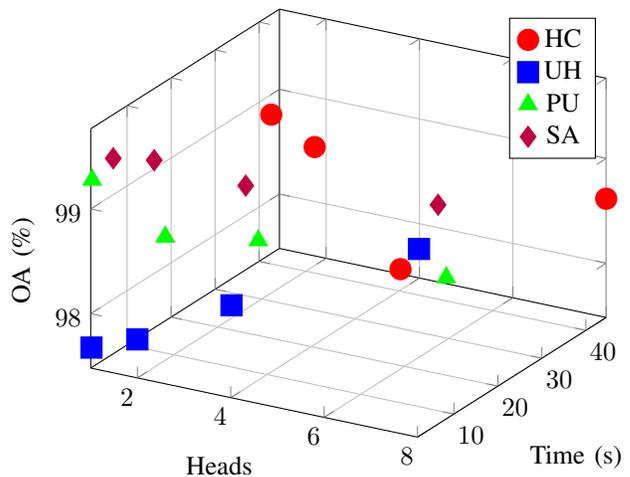

Figure \ref{Fig5} presents the relationship between the number of attention heads, the processing time (in seconds), and the OA for four distinct datasets. The axes of the 3D plot correspond to the number of heads, time, and OA, respectively. This three-dimensional representation facilitates the understanding of the trade-offs between model complexity (as determined by the number of heads) and computational efficiency (measured in time), while simultaneously highlighting the impact of these factors on classification performance. Each dataset is represented by distinct markers and colors: HC (red, circular markers), UH (blue, square markers), PU (green, triangular markers), and SA (purple, diamond markers). The dataset-specific points are plotted for varying values of the "heads" parameter (1, 2, 4, and 8), showing how the OA and time vary with respect to the number of attention heads.

From the figure, several key observations can be made. Firstly, for all datasets, increasing the number of heads from 1 to 8 results in a slight variation in accuracy, suggesting that the model's performance tends to stabilize after a certain point. The trend also reveals that for certain datasets, such as HC and UH, the OA continues to improve slightly even as the number of heads increases, while for others, like PU and SA, the improvement becomes marginal. This indicates that, for some datasets, the model's ability to learn more complex representations with a higher number of attention heads leads to better classification performance, while for others, the added complexity does not significantly enhance the results.

Additionally, the processing time increases with the number of heads for all datasets, reflecting the computational cost associated with a higher number of attention heads. This time increase is generally more pronounced for datasets that require higher computational resources, such as UH and PU, where the time difference between using 1 and 8 heads is more substantial. This highlights the trade-off between the number of heads and time efficiency—while more heads may improve the OA, it may also incur higher computational costs, especially in large-scale data scenarios.

\section{Impact of different Attention methods}

Attention mechanisms have become a cornerstone in modern deep learning models, particularly in the context of HSIC, where capturing intricate spatial and spectral dependencies is critical. Various attention methods, including self and cross-attention, offer different mechanisms for focusing on relevant features in the data. Each approach has its own strengths in emphasizing specific relationships, whether between individual pixels, across different spectral bands, or between input tokens and context. In this section, we explore the impact of different attention methods on \textit{DiffFormer's} performance, assessing their ability to effectively capture complex interactions within HSIs. By comparing these methods across different experimental settings, we aim to determine how each contributes to improving classification accuracy and model robustness, while also considering the trade-offs in computational efficiency and accuracy.

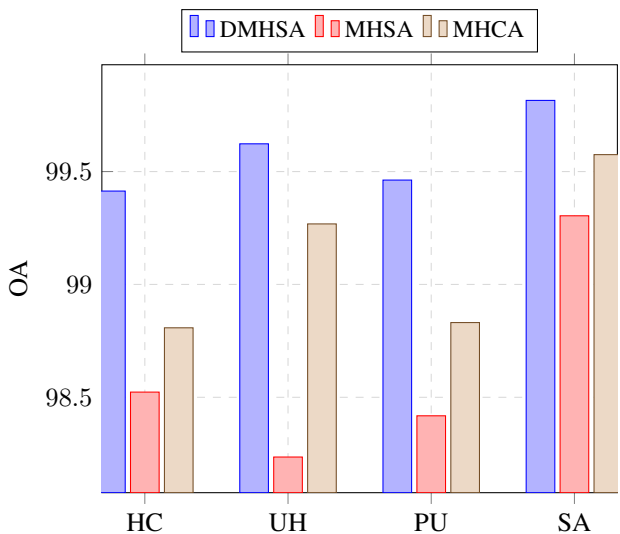
\begin{figure}[!hbt]
    \centering
    \begin{tikzpicture}
        \begin{axis}[
            ybar,
            bar width=0.20, 
            grid=both,
            grid style={dashed, gray!30},
            ylabel={OA},
            xtick={1,2,3,4},
            xticklabels={HC, UH, PU, SA},
            legend style={
                at={(0.5,1.13)},
                anchor=north, 
                font=\small,
                legend columns=3,
                /tikz/every even column/.append style={column sep=0.05cm}},
        ]
            \addplot coordinates {(1,99.4136) (2,99.6229) (3,99.4623)  (4,99.8152)};
            \addplot coordinates {(1,98.5231) (2,98.2355) (3,98.4181)  (4,99.3041)};
            \addplot coordinates {(1,98.8079) (2,99.2681) (3,98.8311)  (4,99.5750)};
            
            \legend{DMHSA, MHSA, MHCA}
        \end{axis}
    \end{tikzpicture}
    \caption{Impact of Attention Mechanisms on HSIC: The bar plot illustrates the Overall Accuracy (OA) achieved by the \textit{DiffFormer} model across three distinct attention mechanisms for four datasets, utilizing a patch size of 12 $\times$ 12, four attentional heads, and four transformer layers.}
    \label{Fig6}
\end{figure}

Figure \ref{Fig6} presents a comparative analysis of the OA achieved by the \textit{DiffFormer} model using three different attention mechanisms: Differential Multi-head Self Attention (DMHSA), Multi-head Self Attention (MHSA), and Multi-head Cross Attention (MHCA). The results are evaluated across four benchmark hyperspectral datasets. DMHSA, the proposed attention mechanism in this work, consistently outperforms the other two methods across all datasets, showcasing its effectiveness in enhancing classification performance.

For the HC dataset, DMHSA achieves the highest OA of 99.41\%, followed by MHCA at 98.81\%, while MHSA lags slightly behind at 98.52\%. A similar trend is observed for the UH dataset, where DMHSA achieves 99.62\%, surpassing MHSA (98.23\%) and MHCA (99.27\%). On the PU dataset, DMHSA again leads with an OA of 99.46\%, compared to 98.83\% by MHCA and 98.41\% by MHSA. For the SA dataset, the performance gap widens further, with DMHSA reaching 99.81\%, outperforming MHCA at 99.57\% and MHSA at 99.30\%.

The superior performance of DMHSA highlights the advantages of introducing differential attention mechanisms, which effectively capture subtle variations in spatial-spectral dependencies across HSIs. In contrast, MHSA and MHCA exhibit slightly lower accuracies, potentially due to their limited ability to differentiate between local and global contextual features. The consistent performance gains observed for DMHSA across diverse datasets validate its robustness and generalizability, making it a promising approach for HSIC.

\section{Comparative Results}

This section provides a detailed discussion on comparative results of proposed \textit{DiffFormer} against several SOTA methods for HSIC to provide a comprehensive performance analysis. The selected comparative methods include a variety of advanced architectures that leverage spatial-spectral information, transformer-based designs, hybrid approaches, and state-space models (Mamba) to address the challenges inherent in HSIC tasks. Specifically, we compare \textit{DiffFormer} with the Attention Graph Convolutional Network (AGCN) \cite{10409250}, the Pyramid Hierarchical Spatial-Spectral Transformer (PyFormer) \cite{10681622}, the Spatial–Spectral Transformer With Conditional Position Encoding (Former) \cite{10604879}, the Spectral–Spatial Wavelet Transformer (WaveFormer) \cite{10399798}, the Hybrid Convolution Transformer (HViT) \cite{2330979}, the Multi-head Spatial-Spectral Mamba (MHSSMamba) \cite{ahmad2024}, and the Spatial-Spectral Wavelet Mamba (WaveMamba) \cite{10767233}.

These methods were evaluated on the four previously mentioned HSI datasets to ensure fair and consistent comparisons. For all methods, we adhered to the experimental settings provided in their respective publications, while uniformly employing a $12 \times 12$ patch size for input data across all datasets. This ensures a standardized framework for assessing classification performance, eliminating biases related to patch size differences. By benchmarking against a diverse set of models, ranging from graph-based approaches to transformer, hybrid, and Mamba architectures, we aim to highlight the strengths and limitations of the proposed \textit{DiffFormer} in comparison with existing methods.

\begin{table}[!hbt]
    \centering
    \caption{Performance comparison on the \textbf{HC dataset} across class-wise accuracies, aggregate metrics, and computational time.}
    \resizebox{\columnwidth}{!}{\begin{tabular}{c|cccccccc} \hline
        \textbf{Class} & AGCN & Former & PyFormer & WaveFormer & HViT & MHMamba & WaveMamba & \textbf{\textit{DiffFormer}} \\ \hline 
        Strawberry & 98.7257 & 99.2548 & 99.5230 & 99.7540 & 99.5305 & 94.9850 & 98.8524 & 99.6199 \\
        Cowpea & 99.1210 & 99.0477 & 99.4872 & 99.0770 & 99.6484 & 90.0673 & 98.1687 & 99.7509 \\
        Soybean & 99.0926 & 99.7731 & 99.7083 & 99.4815 & 99.8379 & 92.2877 & 97.6020 & 99.2546 \\
        Sorghum & 98.6924 & 99.5018 & 98.8792 & 100 & 99.8754 & 93.8978 & 99.2528 & 99.9377 \\
        Water spinach & 91.6666 & 98.8888 & 99.7222 & 99.7222 & 100 & 70.5555 & 97.7777 & 99.7222 \\
        Watermelon & 95.2941 & 96.2500 & 98.0882 & 86.3970 & 92.3529 & 48.1617 & 85.4411 & 94.8529 \\
        Greens & 95.4827 & 98.8706 & 99.7741 & 97.8543 & 98.5883 & 93.6758 & 98.7577 & 98.0237 \\
        Trees & 98.7391 & 98.5722 & 99.7218 & 97.5523 & 97.1815 & 85.9447 & 95.6054 & 99.5735 \\
        Grass & 95.6001 & 99.4720 & 99.7536 & 98.3104 & 98.5568 & 86.5188 & 97.2896 & 98.9792 \\
        Red roof & 99.9049 & 99.8098 & 99.8415 & 99.6830 & 99.9366 & 97.3375 & 99.1759 & 99.4294 \\
        Gray roof & 99.4677 & 99.4874 & 99.7831 & 98.6792 & 99.5466 & 94.1060 & 98.0484 & 99.2903 \\
        Plastic & 97.0108 & 97.6449 & 97.5543 & 89.8550 & 98.3695 & 72.3731 & 98.3695 & 98.9130 \\
        Bare soil & 96.0877 & 96.0877 & 95.8318 & 96.1243 & 95.6855 & 75.6124 & 91.9561 & 96.9287 \\
        Road & 99.6048 & 94.8096 & 99.5689 & 97.2701 & 99.0301 & 91.5229 & 98.3477 & 99.4073 \\
        Bright object & 97.06744 & 97.6539 & 100 & 98.2404 & 98.8269 & 61.5835 & 82.9912 & 98.2404 \\
        Water & 99.8320 & 99.9248 & 99.9734 & 99.9602 & 99.9204 & 99.2528 & 99.7038 & 99.9381 \\ \hline 
        \textbf{$\kappa$} & 98.6568 & 98.4406 & 99.2007 & 97.3726 & 98.5555 & 84.2427 & 97.7480 & 99.8664 \\ \hline
        \textbf{OA} & 98.8519 & 98.7489 & 99.4456 & 98.5176 & 99.0108 & 91.0975 & 98.0753 & 99.3137 \\ \hline
        \textbf{AA} & 97.5868 & 98.9308 & 99.5262 & 98.7341 & 99.1547 & 92.3918 & 96.0837 & 99.4136 \\ \hline
        \textbf{Time (s)} & 2347.30 & 3606.86 & 85926.77 & 3622.55 & 3832.18 & 52495.05 & 36811.62 & 3669.13 \\ \hline
    \end{tabular}}
    \label{Tab3}
\end{table}
\begin{figure*}[!hbt]
    \centering
    \begin{subfigure}{0.11\textwidth}
	\includegraphics[width=0.99\textwidth]{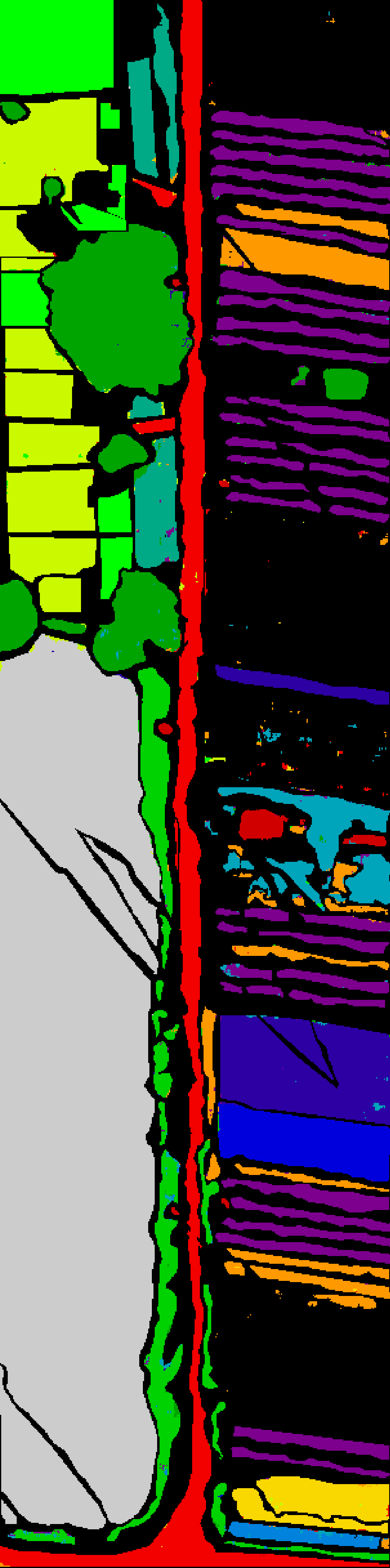}
	\caption*{AGCN} 
    \end{subfigure}
    \begin{subfigure}{0.11\textwidth}
	\includegraphics[width=0.99\textwidth]{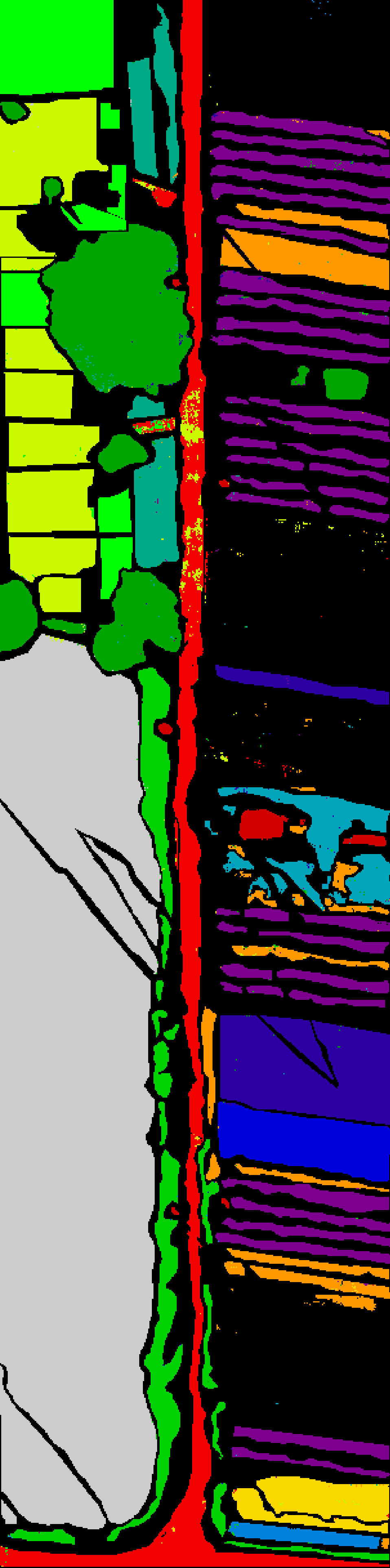}
	\caption*{Former}
    \end{subfigure}
    \begin{subfigure}{0.11\textwidth}
	\includegraphics[width=0.99\textwidth]{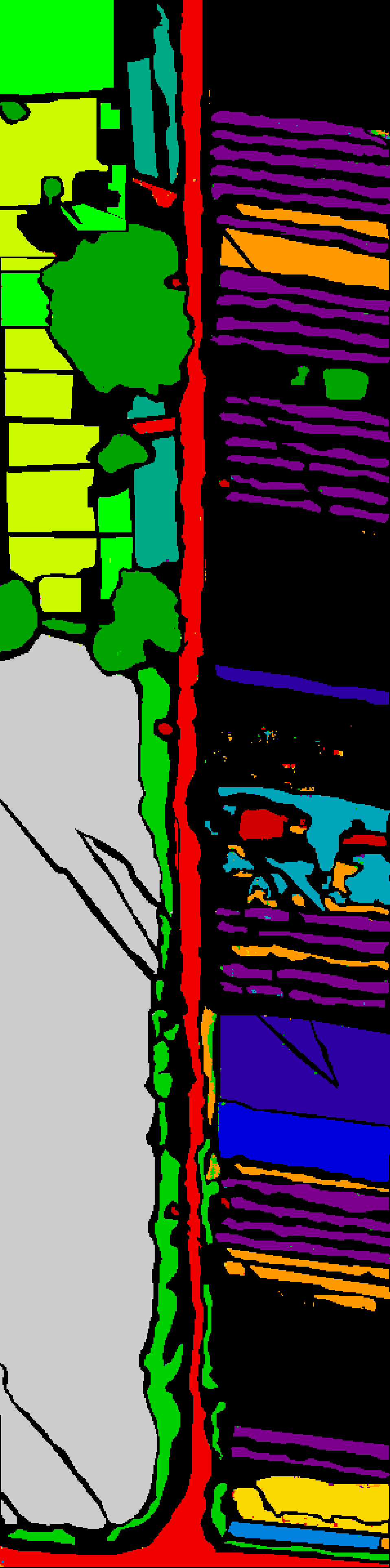}
	\caption*{PyFormer}
    \end{subfigure} 
    \begin{subfigure}{0.11\textwidth}
	\includegraphics[width=0.99\textwidth]{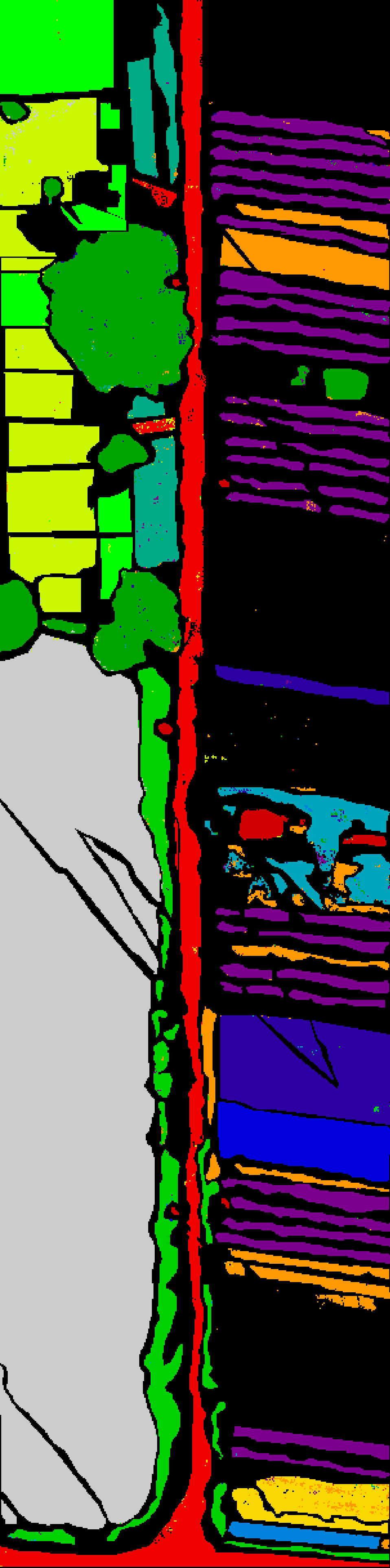}
	\caption*{WaveFormer}
    \end{subfigure}
    \begin{subfigure}{0.11\textwidth}
	\includegraphics[width=0.99\textwidth]{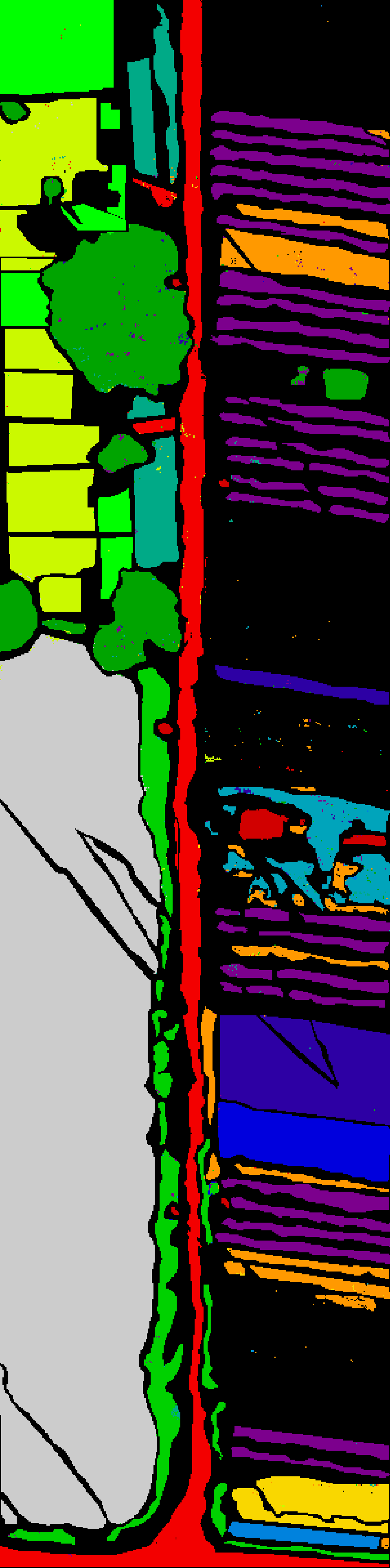}
	\caption*{HViT}
    \end{subfigure}
    \begin{subfigure}{0.11\textwidth}
	\includegraphics[width=0.99\textwidth]{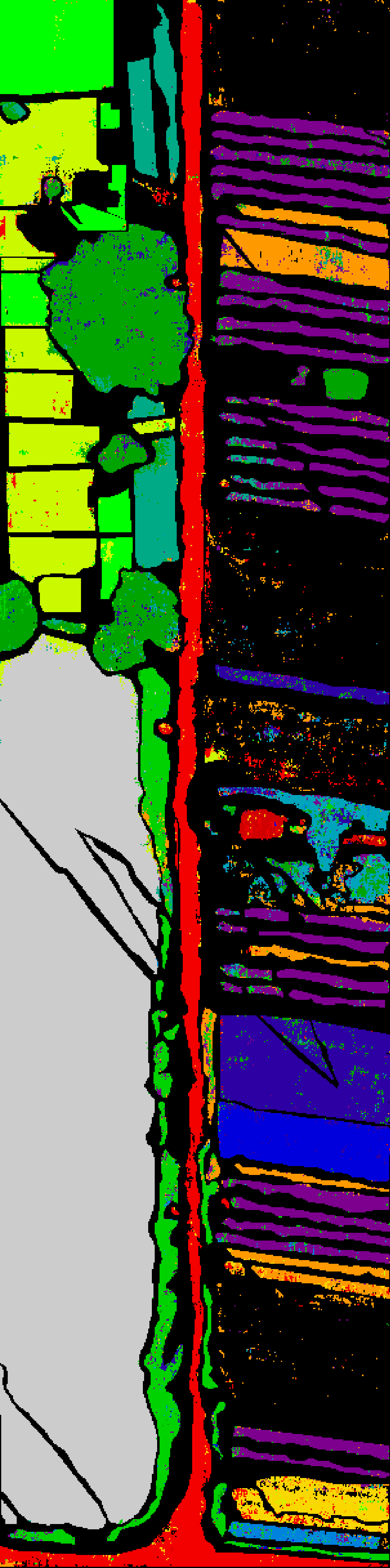}
	\caption*{MHMamba}
    \end{subfigure}
    \begin{subfigure}{0.11\textwidth}
	\includegraphics[width=0.99\textwidth]{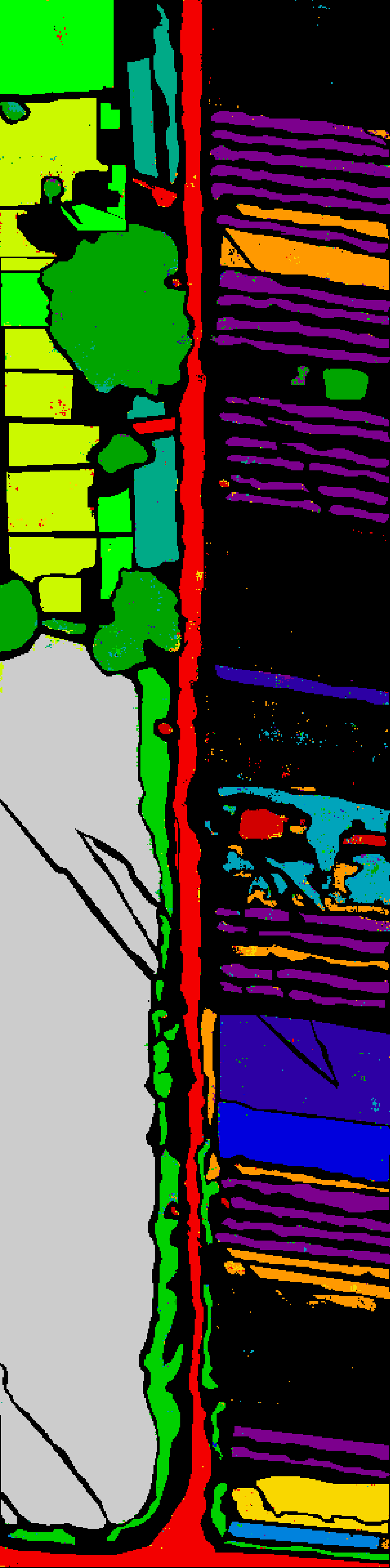}
	\caption*{WaveMamba}
    \end{subfigure}
    \begin{subfigure}{0.11\textwidth}
	\includegraphics[width=0.99\textwidth]{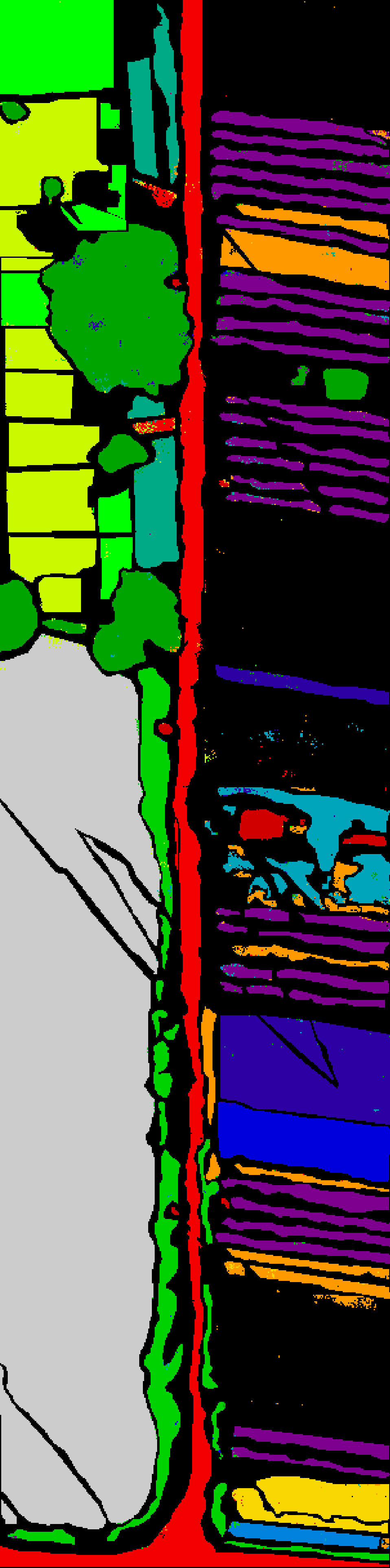}
	\caption*{DiffFormer}
    \end{subfigure}
\caption{Classification maps for the \textbf{HC dataset}, highlighting spatial variability and class-specific performance.}
\label{Fig7}
\end{figure*}
\begin{figure*}[!hbt]
    \centering
    \begin{subfigure}{0.49\textwidth}
	\includegraphics[width=0.99\textwidth]{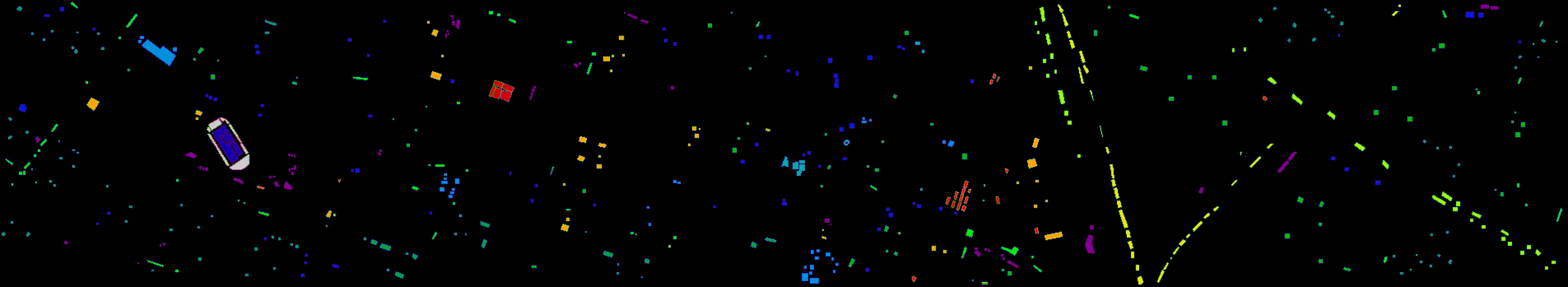}
	\caption*{AGCN} 
    \end{subfigure}
    \begin{subfigure}{0.49\textwidth}
	\includegraphics[width=0.99\textwidth]{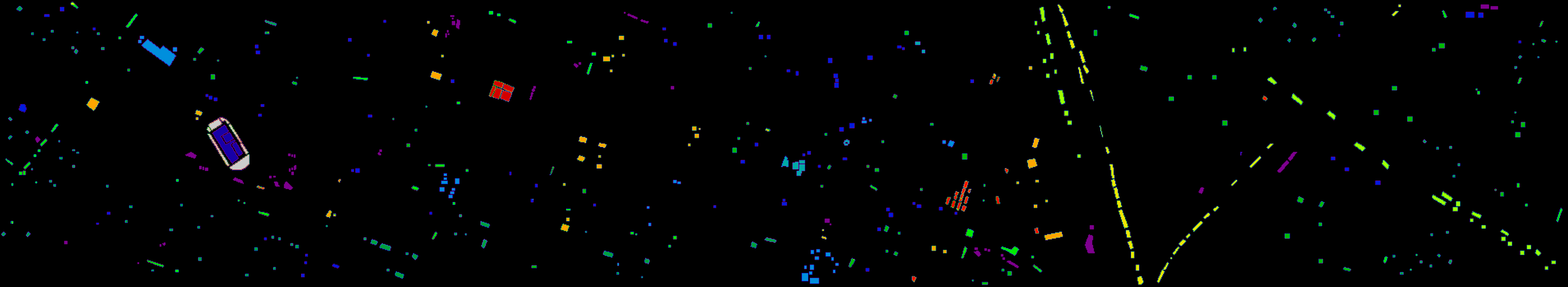}
	\caption*{Former}
    \end{subfigure}
    \begin{subfigure}{0.49\textwidth}
	\includegraphics[width=0.99\textwidth]{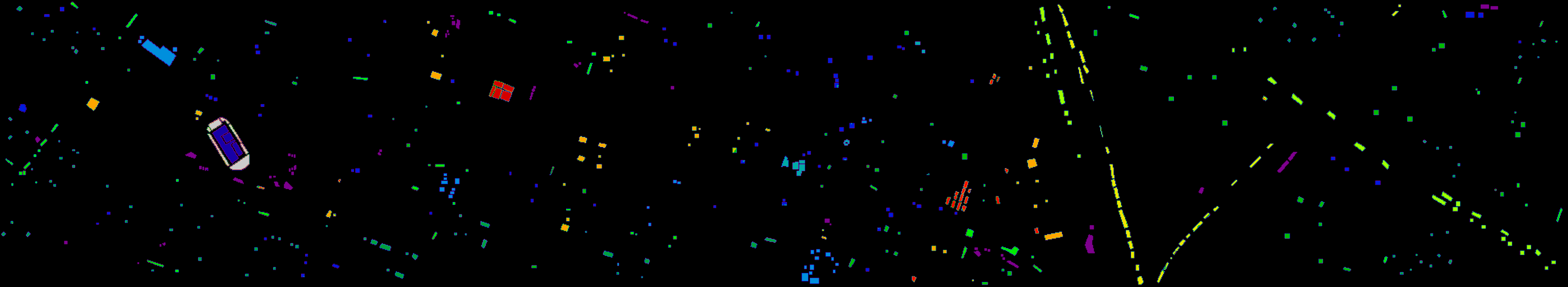}
	\caption*{PyFormer}
    \end{subfigure} 
    \begin{subfigure}{0.49\textwidth}
	\includegraphics[width=0.99\textwidth]{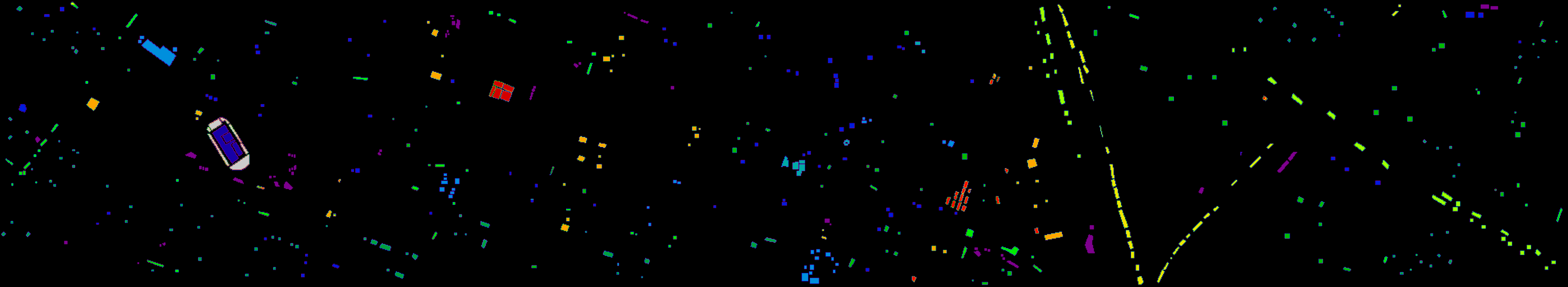}
	\caption*{WaveFormer}
    \end{subfigure}
    \begin{subfigure}{0.49\textwidth}
	\includegraphics[width=0.99\textwidth]{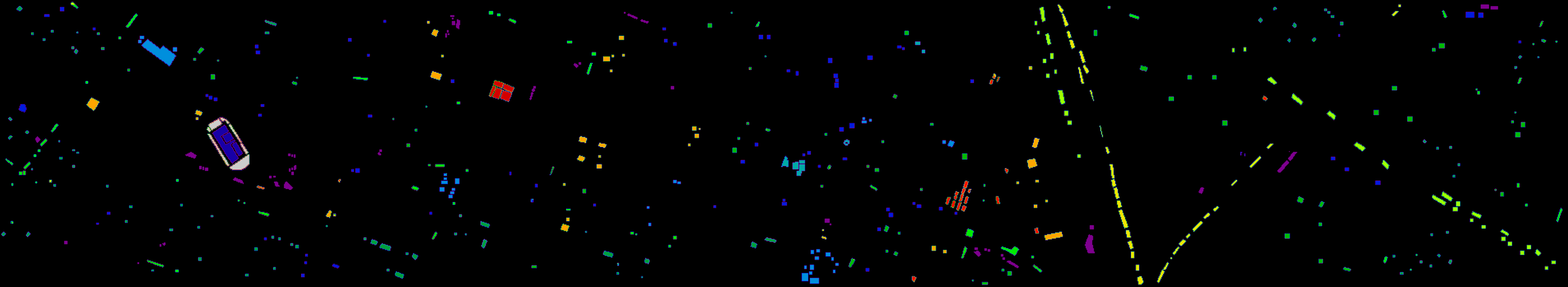}
	\caption*{HViT}
    \end{subfigure}
    \begin{subfigure}{0.49\textwidth}
	\includegraphics[width=0.99\textwidth]{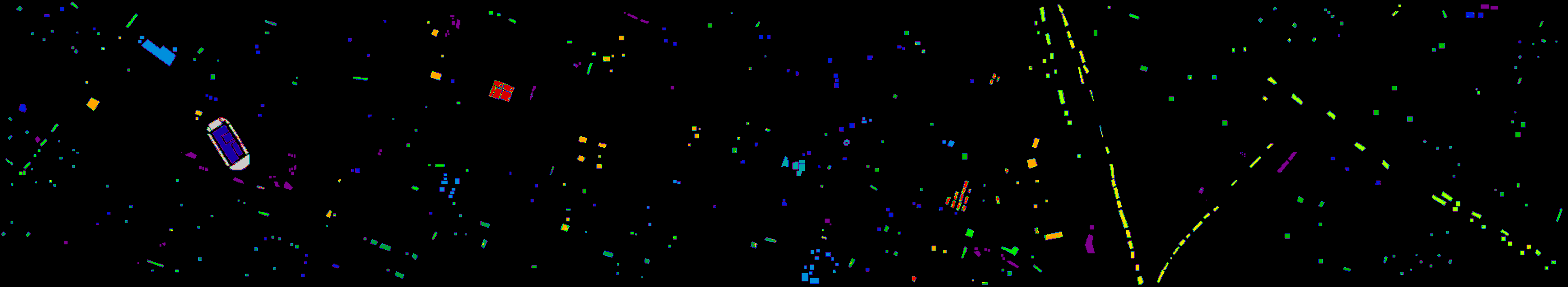}
	\caption*{MHMamba}
    \end{subfigure}
    \begin{subfigure}{0.49\textwidth}
	\includegraphics[width=0.99\textwidth]{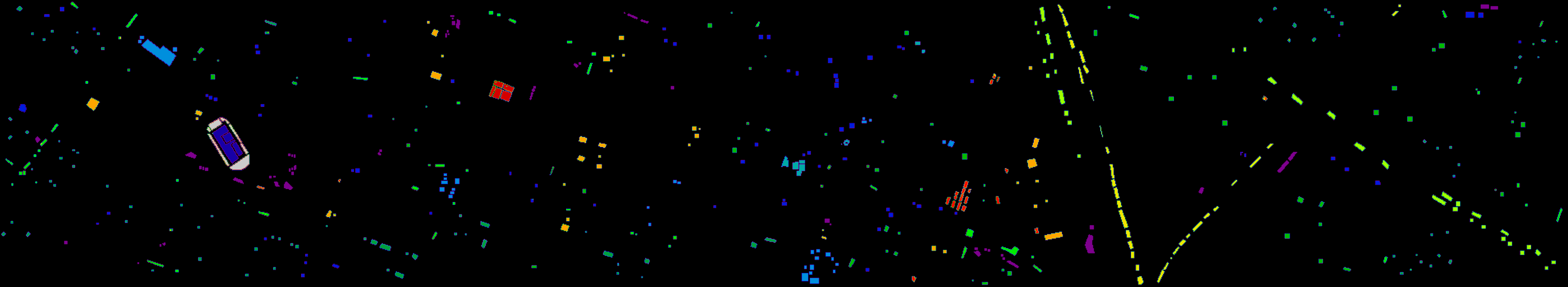}
	\caption*{WaveMamba}
    \end{subfigure}
    \begin{subfigure}{0.49\textwidth}
	\includegraphics[width=0.99\textwidth]{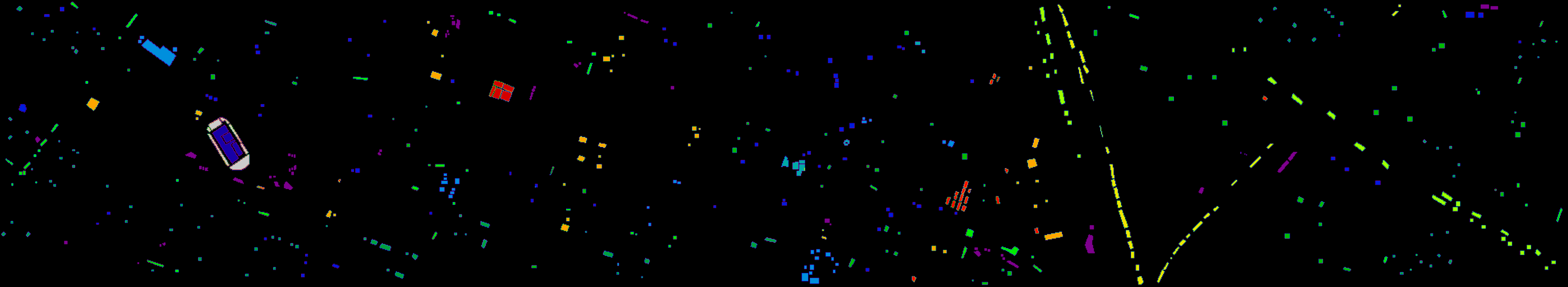}
	\caption*{DiffFormer}
    \end{subfigure}
\caption{Classification maps for the \textbf{UH dataset}, highlighting spatial variability and class-specific performance.}
\label{Fig8}
\end{figure*}

Table \ref{Tab3} provides a comprehensive evaluation of various SOTA models on the HC dataset, reporting per-class accuracy, OA, AA, $\kappa$, and computational time. The results clearly demonstrate that \textit{DiffFormer} consistently outperforms other methods across most metrics, particularly in terms of $\kappa$ and OA, achieving values of 99.8664\% and 99.3137\%, respectively. This indicates the robustness and precision of \textit{DiffFormer}. Notably, DiffFormer achieves high per-class accuracy for critical classes such as Strawberry (99.6199\%), Cowpea (99.7509\%), and Sorghum (99.9377\%), closely matching or exceeding the performance of the best competitors like WaveFormer and HViT. It also demonstrates competitive accuracy for challenging classes like Water spinach (99.7222\%), whereas other models, such as MHMamba, perform significantly worse (70.5555\%). Additionally, DiffFormer maintains computational efficiency with a runtime of 3669.13 seconds, which is significantly lower than models such as PyFormer (85926.77 seconds) and MHMamba (52495.05 seconds), highlighting its suitability for practical applications.

Figure \ref{Fig7} visually depicts the classification maps generated by the evaluated models. The qualitative results further corroborate the quantitative findings from Table \ref{Tab3}, showing that \textit{DiffFormer} produces more accurate and spatially coherent classification maps compared to its counterparts. Models like MHMamba and WaveMamba exhibit noticeable misclassifications and noisy outputs, particularly in heterogeneous regions. In contrast, \textit{DiffFormer} achieves smooth and precise segmentation, effectively capturing spatial-spectral relationships in the data.

\begin{table}[!hbt]
    \centering
    \caption{Performance comparison on the \textbf{UH dataset} across class-wise accuracies, aggregate metrics, and computational time.}
    \resizebox{\columnwidth}{!}{\begin{tabular}{c|cccccccc} \hline
        \textbf{Class} & AGCN & Former & PyFormer & WaveFormer & HViT & MHMamba & WaveMamba & \textbf{\textit{DiffFormer}} \\ \hline 

        Healthy grass & 99.3610 & 99.5207 & 100 & 99.5207 & 98.8817 & 99.0415 & 98.4025 & 98.9333 \\
        Stressed grass & 99.8405 & 99.8405 & 99.5215 & 99.8405 & 99.8405 & 97.7671 & 99.8405 & 99.2021 \\
        Synthetic grass & 99.7126 & 100 & 100 & 100 & 100 & 99.7126 & 99.7126 & 100 \\
        Trees & 99.8392 & 100 & 98.8745 & 99.8392 & 99.6784 & 96.3022 & 99.6784 & 100 \\
        Soil & 99.6779 & 100 & 99.8389 & 100 & 99.8389 & 99.1948 & 100 & 100 \\
        Water & 98.7654 & 100 & 100 & 100 & 98.7654 & 96.2962 & 99.3827 & 96.9387 \\
        Residential & 98.8958 & 97.9495 & 98.4227 & 98.4227 & 97.4763 & 84.7003 & 96.6876 & 99.2105 \\
        Commercial & 95.9807 & 99.0353 & 95.0160 & 99.8392 & 99.5176 & 93.5691 & 99.3569 & 100 \\
        Road & 99.3610 & 99.0415 & 99.8402 & 99.0415 & 99.0415 & 92.6517 & 98.4025 & 99.7340 \\
        Highway & 100 & 100 & 100 & 100 & 100 & 88.4364 & 99.5114 & 100 \\
        Railway & 99.6763 & 100 & 94.8220 & 100 & 99.6763 & 96.7637 & 99.8381 & 100 \\
        Parking Lot 1 & 99.5137 & 99.1896 & 98.5413 & 99.3517 & 99.1896 & 91.7341 & 99.3517 & 100 \\
        Parking Lot 2 & 99.1452 & 91.4529 & 93.1623 & 91.8803 & 94.4444 & 62.8205 & 94.4444 & 97.8723 \\
        Tennis Court & 100 & 99.5327 & 100 & 99.5327 & 100 & 96.7289 & 99.0654 & 100 \\
        Running Track & 100 & 100 & 100 & 100 & 100 & 100 & 100 & 100 \\ \hline 
        \textbf{$\kappa$} & 99.2231 & 99.2086 & 98.4174 & 99.3381 & 99.1655 & 93.1787 & 98.9498 & 99.5923 \\ \hline
        \textbf{OA} & 99.2814 & 99.2681 & 98.5362 & 99.3878 & 99.2282 & 93.6926 & 99.0286 & 99.6229 \\ \hline
        \textbf{AA} & 99.3179 & 99.0375 & 98.5359 & 99.1512 & 99.0900 & 93.0479 & 98.9116 & 99.4594 \\ \hline
        \textbf{Time (s)} & 98.86 & 6392.95 & 1001.73 & 146.79 & 152.05 & 362.30 & 1374.47 & 219.45 \\ \hline
    \end{tabular}}
    \label{Tab4}
\end{table}

Table \ref{Tab4} summarizes the performance across various land cover classes in the UH dataset. Moreover, Figure \ref{Fig8} illustrates the classification maps produced by each model, highlighting spatial variability and class-specific performance. The results show that the proposed \textit{DiffFormer} achieves competitive accuracy across most classes, outperforming other models in critical categories. For example, \textbf{DiffFormer} attains a perfect accuracy (100\%) for high-precision classes such as Synthetic Grass, Highway, and Running Track, demonstrating its robustness in distinguishing spectrally and spatially homogenous regions. Notably, the model also achieves superior performance in challenging categories like Commercial (100\%) and Residential (99.2105\%), where other models show variability. In contrast, models like MHMamba underperform in several categories, such as Parking Lot 2 (62.8205\%) and Water (96.2962\%), indicating its limitations in handling complex spectral-spatial variations. Similarly, HViT exhibits consistent but slightly lower performance across most classes, particularly for Road (99.0415\%) and Tennis Court (100\%). From an aggregate perspective, \textit{DiffFormer} outperforms all competing models, achieving the highest OA (99.6229\%), AA (99.4594\%), and $\kappa$ (99.5923\%). These metrics validate the model's ability to generalize across diverse land cover classes. The second-best model, WaveFormer, delivers comparable performance but falls slightly short in terms of AA (99.1512\%) and OA (99.3878\%), underscoring the improvements introduced by \textit{DiffFormer}.

In terms of computational efficiency, \textit{DiffFormer} exhibits a balanced trade-off between accuracy and time, with a processing time of 219.45 seconds. While WaveFormer is marginally faster (146.79 seconds), its slightly lower performance highlights the significance of the proposed enhancements in DiffFormer. Notably, Former requires the longest processing time (6392.95 seconds), making it impractical for large-scale applications despite its high accuracy. The classification maps in Figure \ref{Fig8} provide a visual validation of the quantitative results. \textit{DiffFormer} exhibits superior spatial coherence and reduced misclassifications, particularly in heterogeneous regions such as Residential and Commercial. Conversely, models like MHMamba and AGCN produce noisier outputs with noticeable artifacts in areas like Soil and Parking Lot 2.

\begin{table}[!hbt]
    \centering
    \caption{Performance comparison on the \textbf{SA dataset} across class-wise accuracies, aggregate metrics, and computational time.}
    \resizebox{\columnwidth}{!}{\begin{tabular}{c|cccccccc} \hline
        \textbf{Class} & AGCN & Former & PyFormer & WaveFormer & HViT & MHMamba & WaveMamba & \textbf{\textit{DiffFormer}} \\ \hline 
        Broccoli 1 & 100 & 100 & 100 & 100 & 100 & 99.8007 & 100 & 100 \\
        Broccoli 2 & 100 & 100 & 100 & 100 & 100 & 100 & 100 & 100 \\
        Fallow & 100 & 100 & 100 & 100 & 100 & 99.2914 & 99.8987 & 100 \\
        Fallow Rough & 99.7130 & 99.8565 & 100 & 99.7130 & 99.8565 & 99.5695 & 100 & 99.5215 \\
        Fallow Smooth & 100 & 99.9253 & 100 & 99.9253 & 99.7012 & 98.0582 & 99.1038 & 99.0037 \\
        Stubble & 100 & 100 & 100 & 100 & 100 & 100 & 99.9494 & 100 \\
        Celery & 99.9441 & 100 & 100 & 100 & 100 & 99.7206 & 99.9441 & 100 \\
        Grapes & 99.4854 & 99.0951 & 99.5919 & 99.1838 & 99.0596 & 97.1965 & 99.2193 & 99.8521 \\
        Soil Vinyard & 99.9677 & 100 & 100 & 100 & 100 & 100 & 100 & 99.8925 \\
        Corn Senesced & 99.7559 & 99.9389 & 99.8779 & 99.8779 & 99.6949 & 99.3288 & 99.9389 & 100 \\
        Lettuce 4wk & 99.4382 & 100 & 98.8764 & 100 & 100 & 99.6254 & 99.8127 & 100 \\
        Lettuce 5wk & 99.8961 & 100 & 100 & 99.8961 & 100 & 100 & 100 & 100 \\
        Lettuce 6wk & 100 & 100 & 100 & 100 & 100 & 99.3449 & 100 & 100 \\
        Lettuce 7wk & 99.4392 & 99.6261 & 99.6261 & 100 & 100 & 98.5046 & 99.8130 & 100 \\
        Vinyard Untrained & 99.4496 & 98.4865 & 90.6989 & 98.2388 & 98.2663 & 73.6378 & 98.4865 & 99.4039 \\
        Vinyard Vertical & 98.7818 & 100 & 99.7785 & 100 & 100 & 99.3355 & 100 & 100 \\ \hline 
        \textbf{$\kappa$} & 99.6914 & 99.5433 & 98.4639 & 99.5227 & 99.4815 & 1477.32 & 99.5186 & 99.7942 \\ \hline
        \textbf{OA} & 99.7228 & 99.5898 & 98.6218 & 99.5714 & 99.5344 & 95.0961 & 99.5677 & 99.8152 \\ \hline
        \textbf{AA} & 99.7419 & 99.8080 & 99.2781 & 99.8021 & 99.7861 & 95.6068 & 99.7604 & 99.8546 \\ \hline
        \textbf{Time (s)} & 328.86 & 561.87 & 3590.99 & 529.59 & 533.54 & 97.7134 & 8526.14 & 767.71 \\ \hline
    \end{tabular}}
    \label{Tab6}
\end{table}
\begin{figure}[!hbt]
    \centering
    \begin{subfigure}{0.11\textwidth}
	\includegraphics[width=0.99\textwidth]{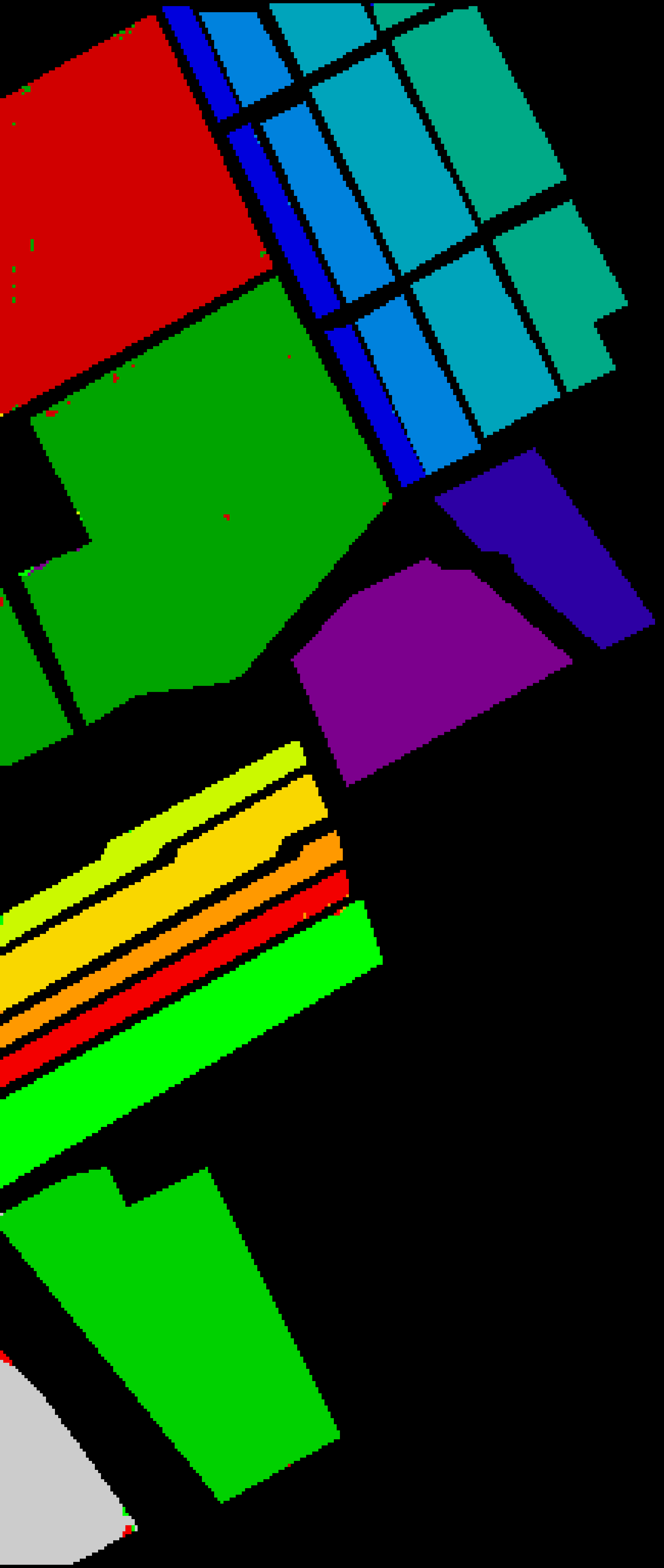}
	\caption*{AGCN} 
    \end{subfigure}
    \begin{subfigure}{0.11\textwidth}
	\includegraphics[width=0.99\textwidth]{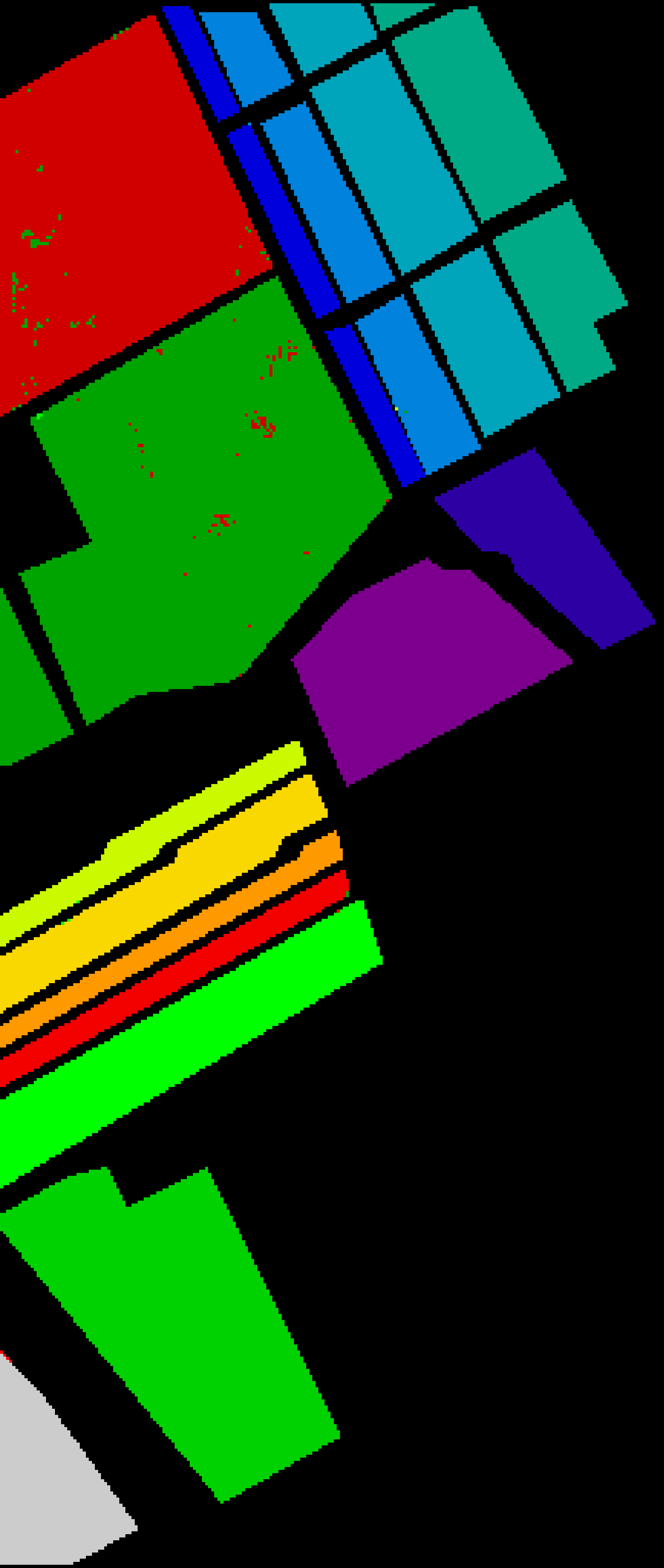}
	\caption*{Former}
    \end{subfigure}
    \begin{subfigure}{0.11\textwidth}
	\includegraphics[width=0.99\textwidth]{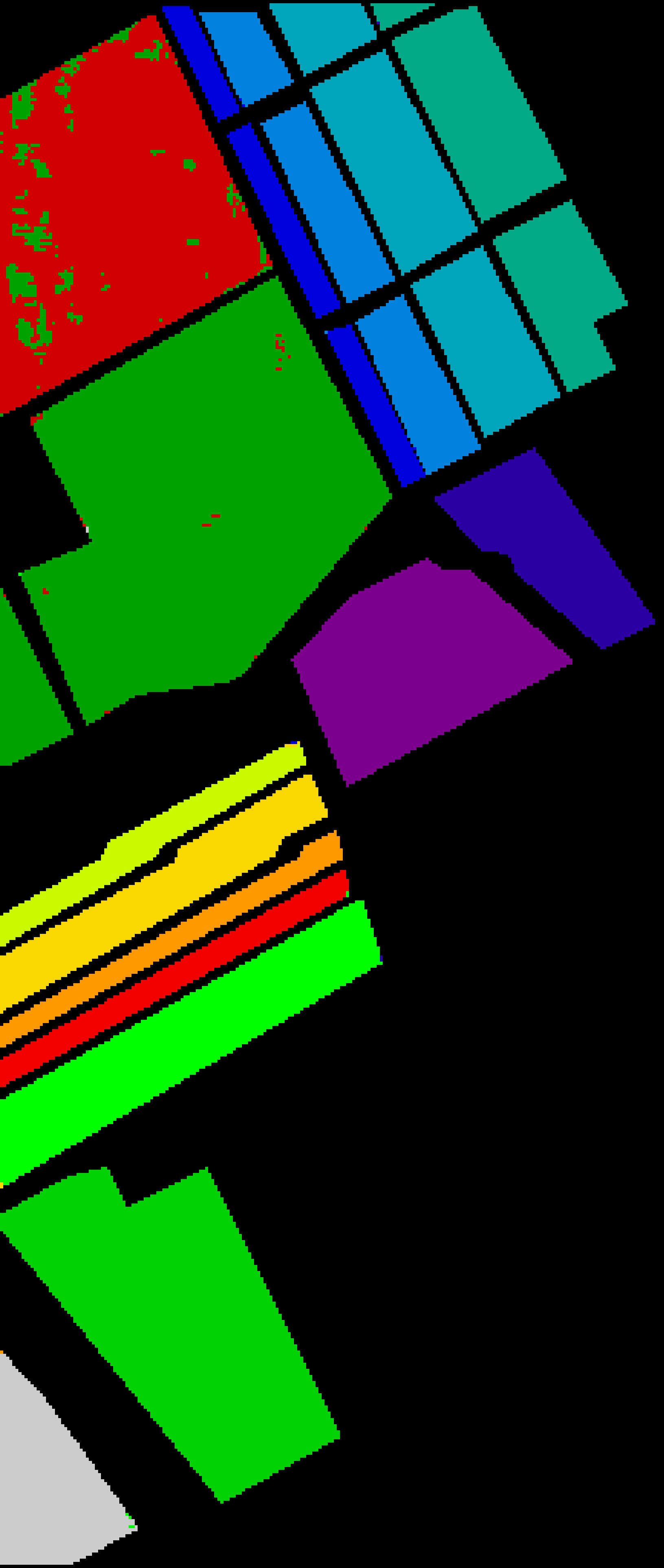}
	\caption*{PyFormer}
    \end{subfigure} 
    \begin{subfigure}{0.11\textwidth}
	\includegraphics[width=0.99\textwidth]{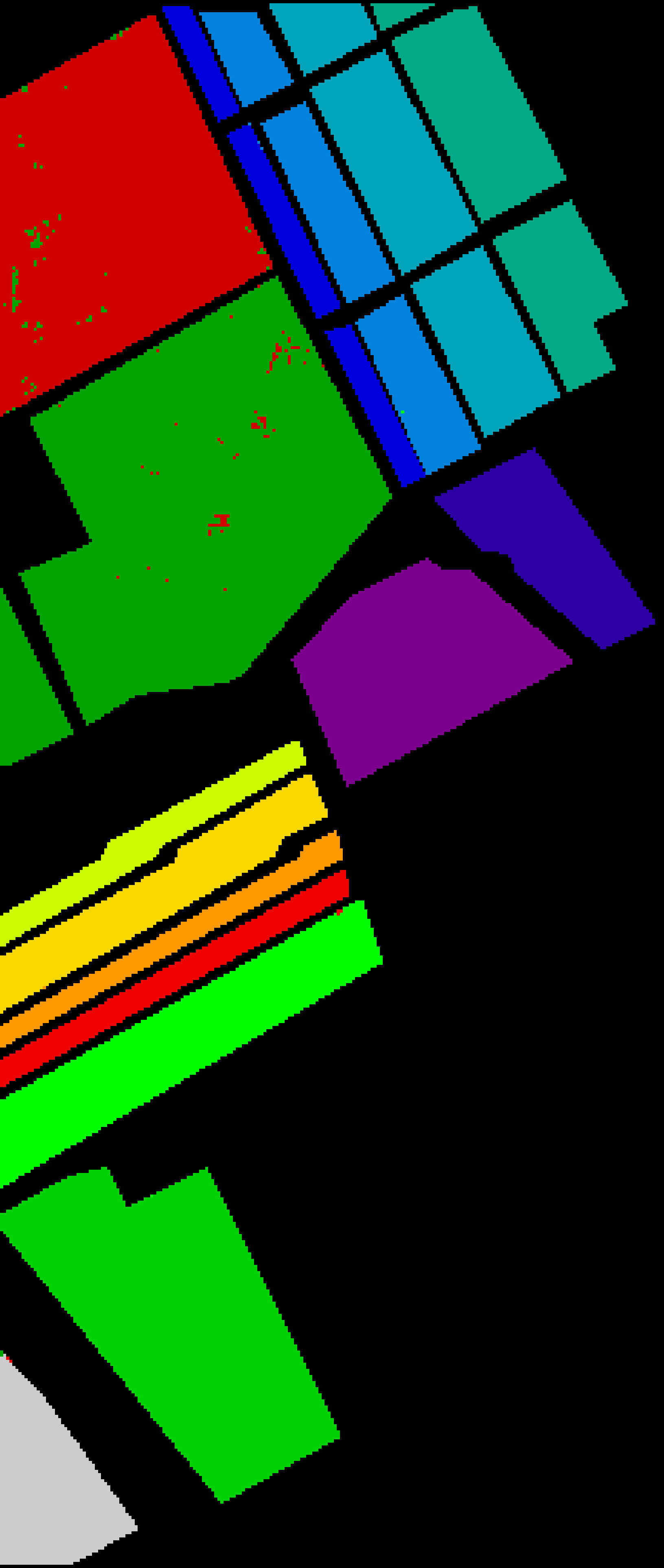}
	\caption*{WaveFormer}
    \end{subfigure}
    \begin{subfigure}{0.11\textwidth}
	\includegraphics[width=0.99\textwidth]{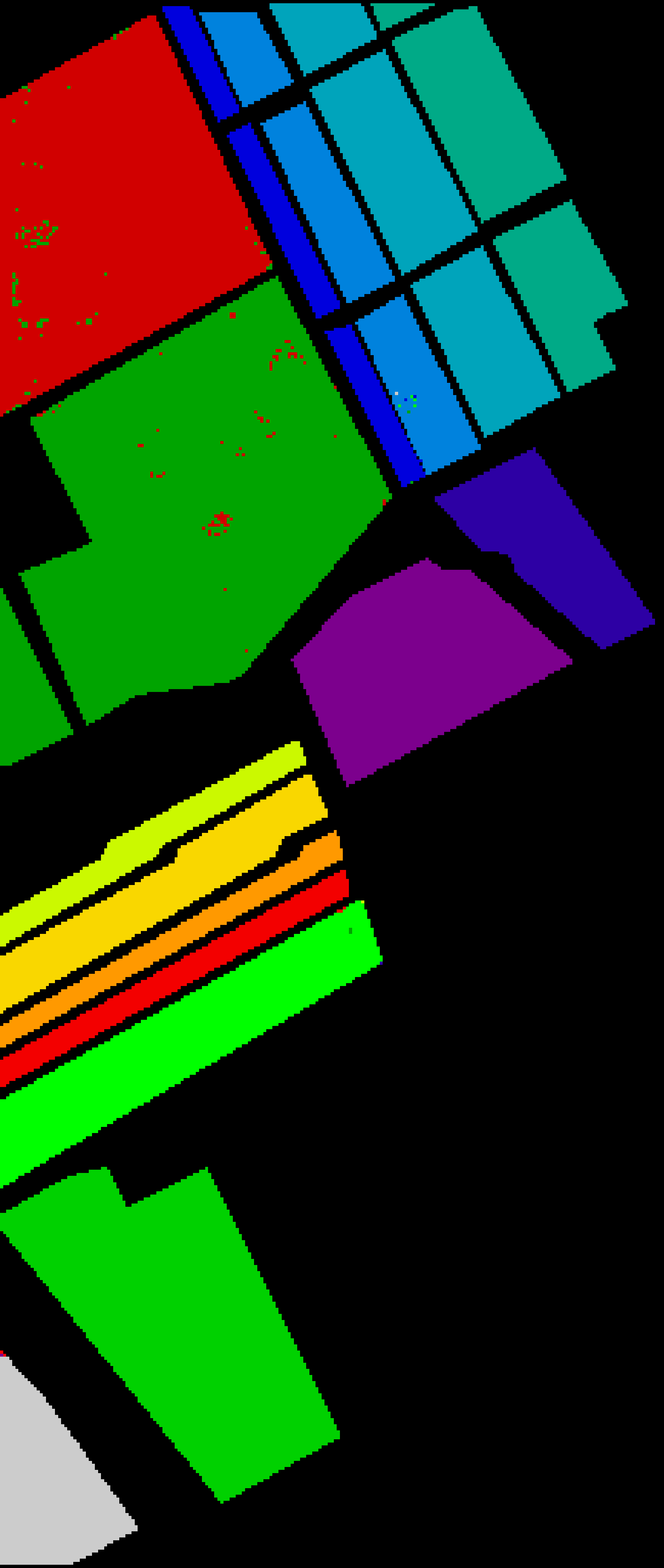}
	\caption*{HViT}
    \end{subfigure}
    \begin{subfigure}{0.11\textwidth}
	\includegraphics[width=0.99\textwidth]{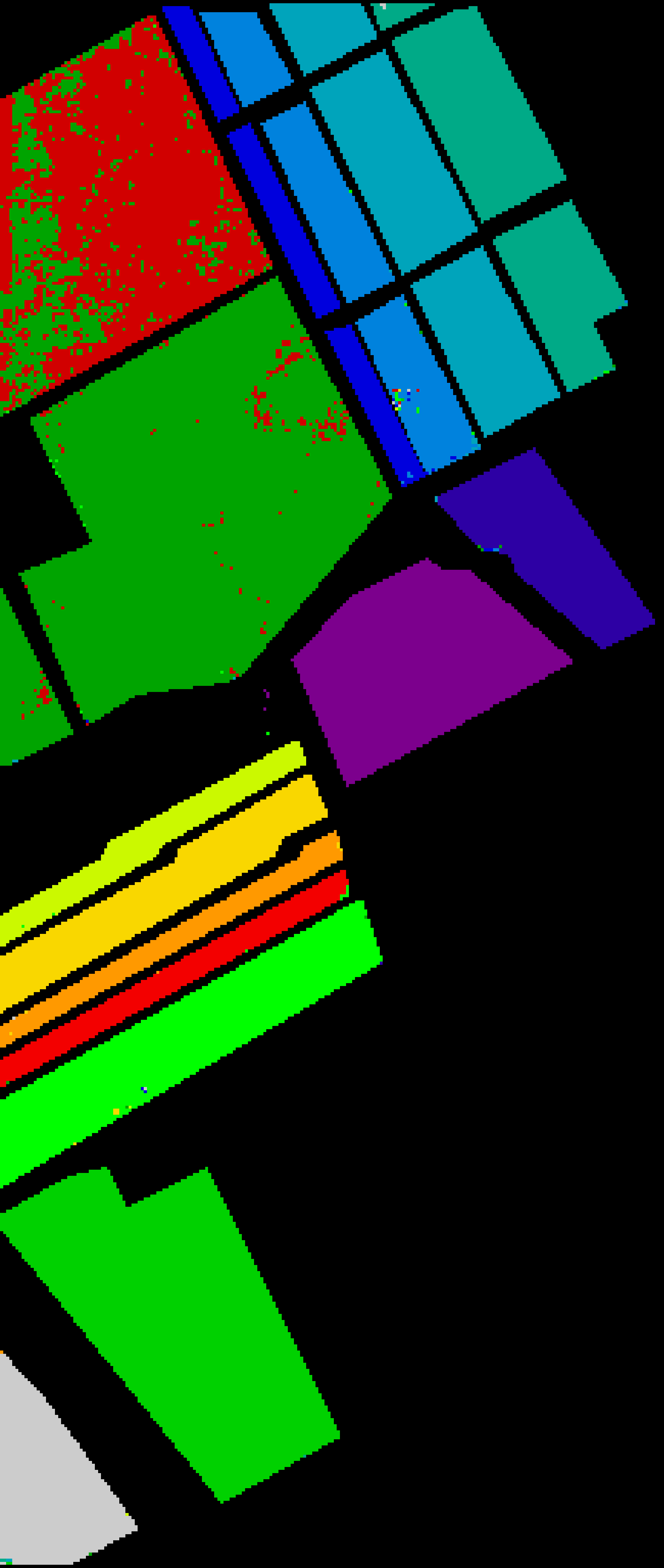}
	\caption*{MHMamba}
    \end{subfigure}
    \begin{subfigure}{0.11\textwidth}
	\includegraphics[width=0.99\textwidth]{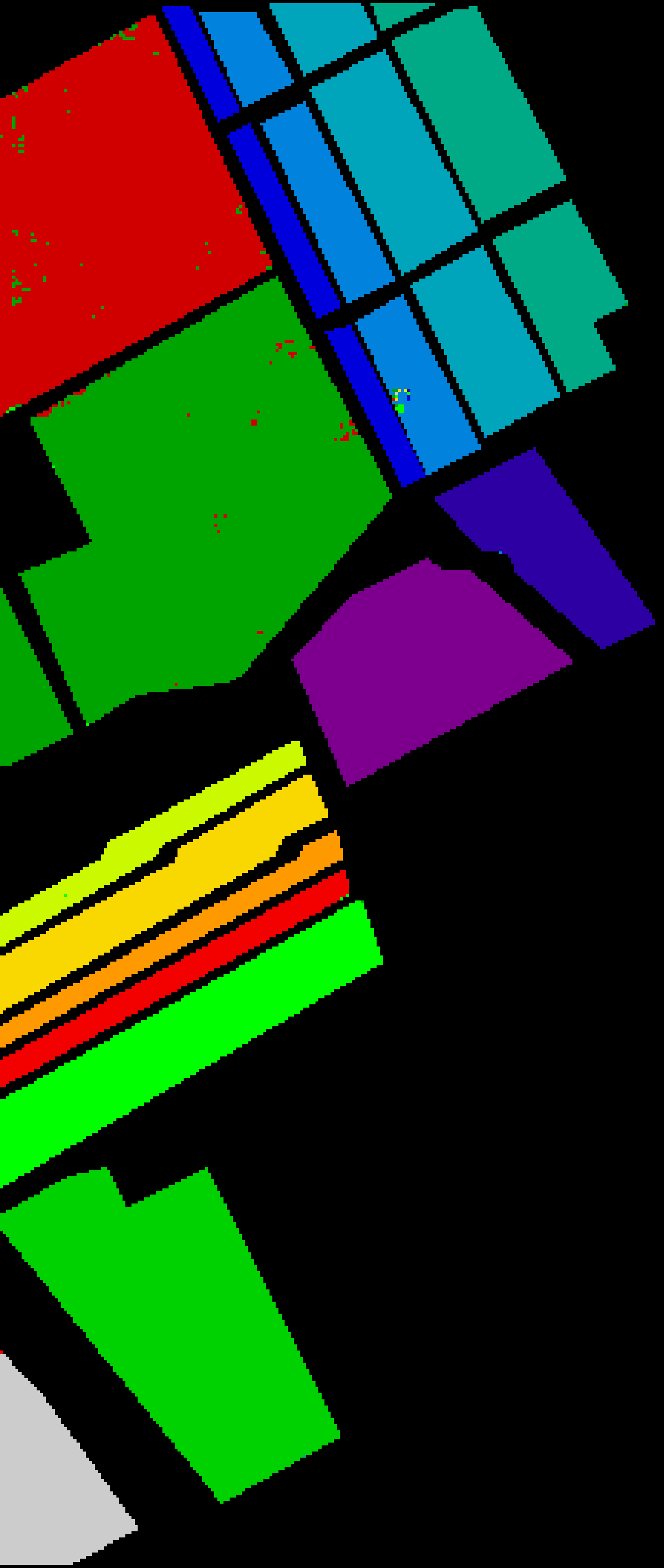}
	\caption*{WaveMamba}
    \end{subfigure}
    \begin{subfigure}{0.11\textwidth}
	\includegraphics[width=0.99\textwidth]{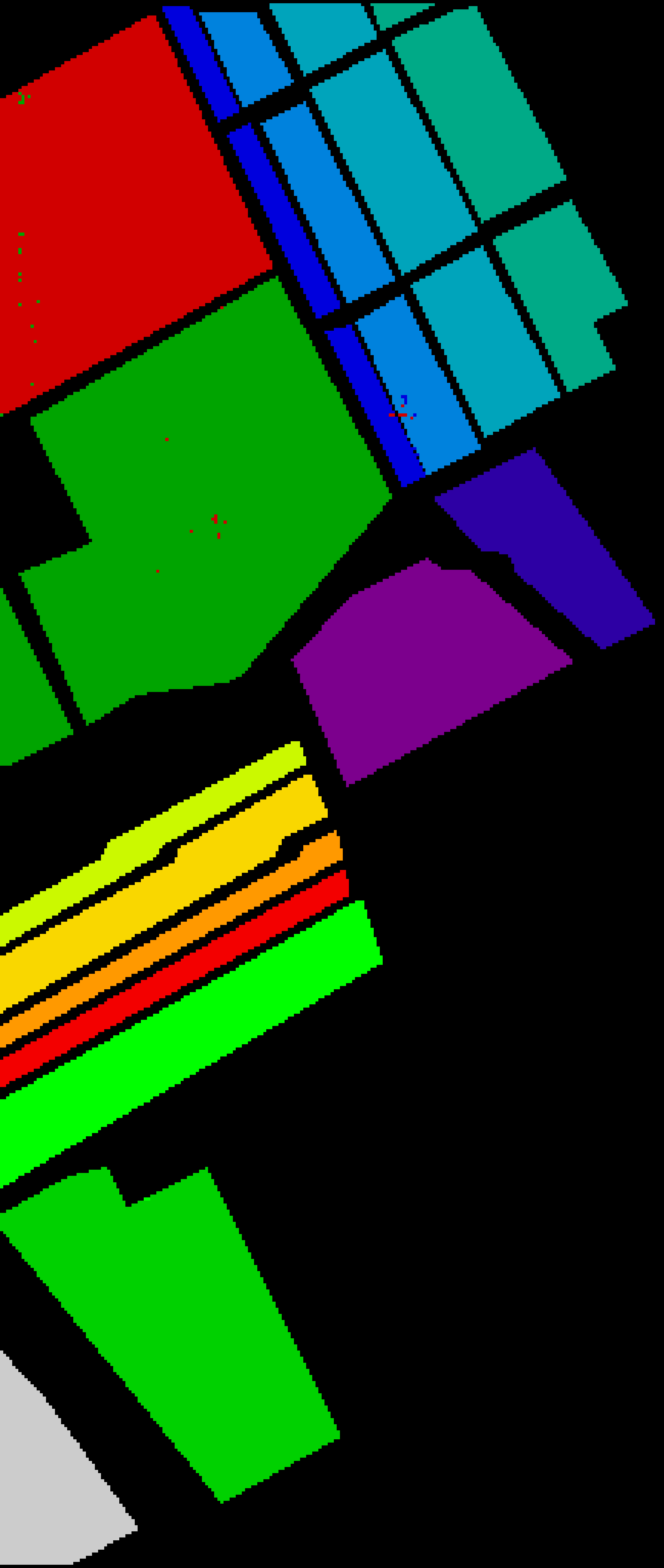}
	\caption*{DiffFormer}
    \end{subfigure}
\caption{Classification maps for the \textbf{SA dataset}, highlighting spatial variability and class-specific performance.}
\label{Fig9}
\end{figure}

The results in Table \ref{Tab6} and Figure \ref{Fig9} comprehensively evaluate and compare the performance of various models on the SA dataset. Table \ref{Tab6} presents class-wise accuracies, meanwhile, Figure \ref{Fig9} displays the corresponding classification maps to visually interpret the spatial and class-specific performance of each model. From Table \ref{Tab6}, the proposed model, \textit{DiffFormer}, outperforms other models in most aggregate metrics, achieving the highest OA (99.8152\%), AA (99.8546\%), and $\kappa$ (99.7942\%). These results indicate superior classification reliability and OA. Class-wise accuracies reveal that \textit{DiffFormer} maintains consistent performance across classes, with perfect scores for challenging categories like "Broccoli 1," "Broccoli 2," and "Stubble," outperforming models like MHMamba, which struggles particularly in "Vinyard Untrained" with a significantly lower accuracy (73.6378\%). Furthermore, \textit{DiffFormer} demonstrates competitive computational efficiency with a runtime of 767.71 seconds, which, while not the fastest, remains practical compared to WaveMamba's extensive runtime of 8526.14 seconds.

Figure \ref{Fig9} provides a qualitative comparison of classification maps generated by the models. It highlights the spatial consistency and accuracy of \textit{DiffFormer}, especially in regions with complex class distributions, such as "Fallow Rough" and "Vinyard Vertical." Misclassifications, evident in other models like PyFormer and MHMamba, are significantly reduced in \textit{DiffFormer}, leading to smoother and more accurate spatial patterns. Additionally, the high fidelity of the maps corroborates the quantitative superiority of \textit{DiffFormer}. The analysis of Table \ref{Tab6} and Figure \ref{Fig9} underscores the effectiveness of \textit{DiffFormer} in balancing classification accuracy and computational efficiency. While other models demonstrate competitive performance in specific metrics or classes, \textit{DiffFormer} provides a well-rounded solution with enhanced accuracy and reliable spatial representation. 

\begin{table}[!hbt]
    \centering
    \caption{Performance comparison on the \textbf{PU dataset} across class-wise accuracies, aggregate metrics, and computational time.}
    \resizebox{\columnwidth}{!}{\begin{tabular}{c|cccccccc} \hline
        \textbf{Class} & AGCN & Former & PyFormer & WaveFormer & HViT & MHMamba & WaveMamba & \textbf{\textit{DiffFormer}} \\ \hline 
        Asphalt & 100 & 99.0349 & 99.8190 & 98.8540 & 98.6429 & 92.8226 & 98.9143 & 99.4972 \\
        Meadows & 99.9678 & 99.9249 & 99.9678 & 99.9249 & 99.9356 & 98.6593 & 100 & 99.9642 \\
        Gravel & 98.2840 & 93.6129 & 97.7121 & 93.8036 & 92.6596 & 83.9847 & 95.2335 & 96.5079 \\
        Trees & 99.3472 & 99.5430 & 99.6736 & 99.2819 & 99.2167 & 92.8198 & 99.2167 & 99.2383 \\
        Metal Sheets & 100 & 100 & 99.8514 & 100 & 100 & 99.7028 & 100 & 100 \\
        Bare Soil & 99.9204 & 99.9602 & 99.6022 & 99.7613 & 99.7215 & 81.7422 & 100 & 99.7349 \\
        Bitumen & 99.8496 & 99.2481 & 99.5488 & 99.0977 & 99.0977 & 81.8045 & 98.7969 & 99.7493 \\
        Bricks & 99.4024 & 96.4149 & 95.7088 & 96.1434 & 95.5458 & 74.7963 & 96.4149 & 98.0090 \\
        Shadows & 98.7341 & 98.9451 & 100 & 98.3122 & 97.8902 & 94.5147 & 98.5232 & 99.6478 \\ \hline 
        \textbf{$\kappa$} & 99.6839 & 98.8229 & 99.0982 & 98.3532 & 98.0789 & 89.2972 & 98.8910 & 99.2872 \\ \hline
        \textbf{OA} & 99.7615 & 99.1116 & 99.1882 & 98.6865 & 98.4816 & 91.9908 & 99.1630 & 99.4623 \\ \hline
        \textbf{AA} & 99.5006 & 98.5204 & 99.3875 & 99.0087 & 98.8544 & 88.9830 & 98.5666 & 99.1498 \\ \hline
        \textbf{Time (s)} & 261.21 & 383.60 & 2840.00 & 383.09 & 405.36 & 1041.99 & 3952.37 & 601.86 \\ \hline
    \end{tabular}}
    \label{Tab5}
\end{table}
\begin{figure}[!hbt]
    \centering
    \begin{subfigure}{0.11\textwidth}
	\includegraphics[width=0.99\textwidth]{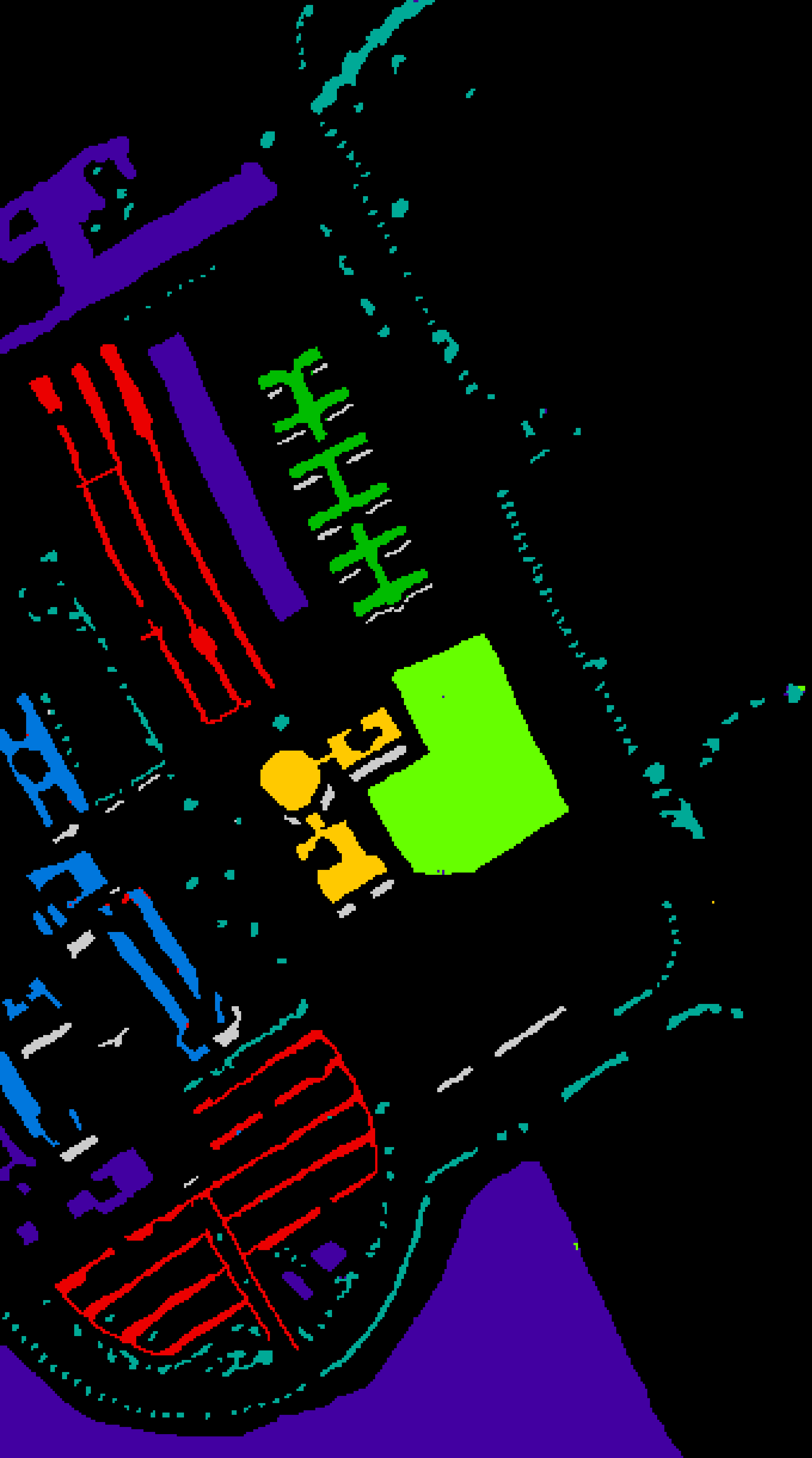}
	\caption*{AGCN} 
    \end{subfigure}
    \begin{subfigure}{0.11\textwidth}
	\includegraphics[width=0.99\textwidth]{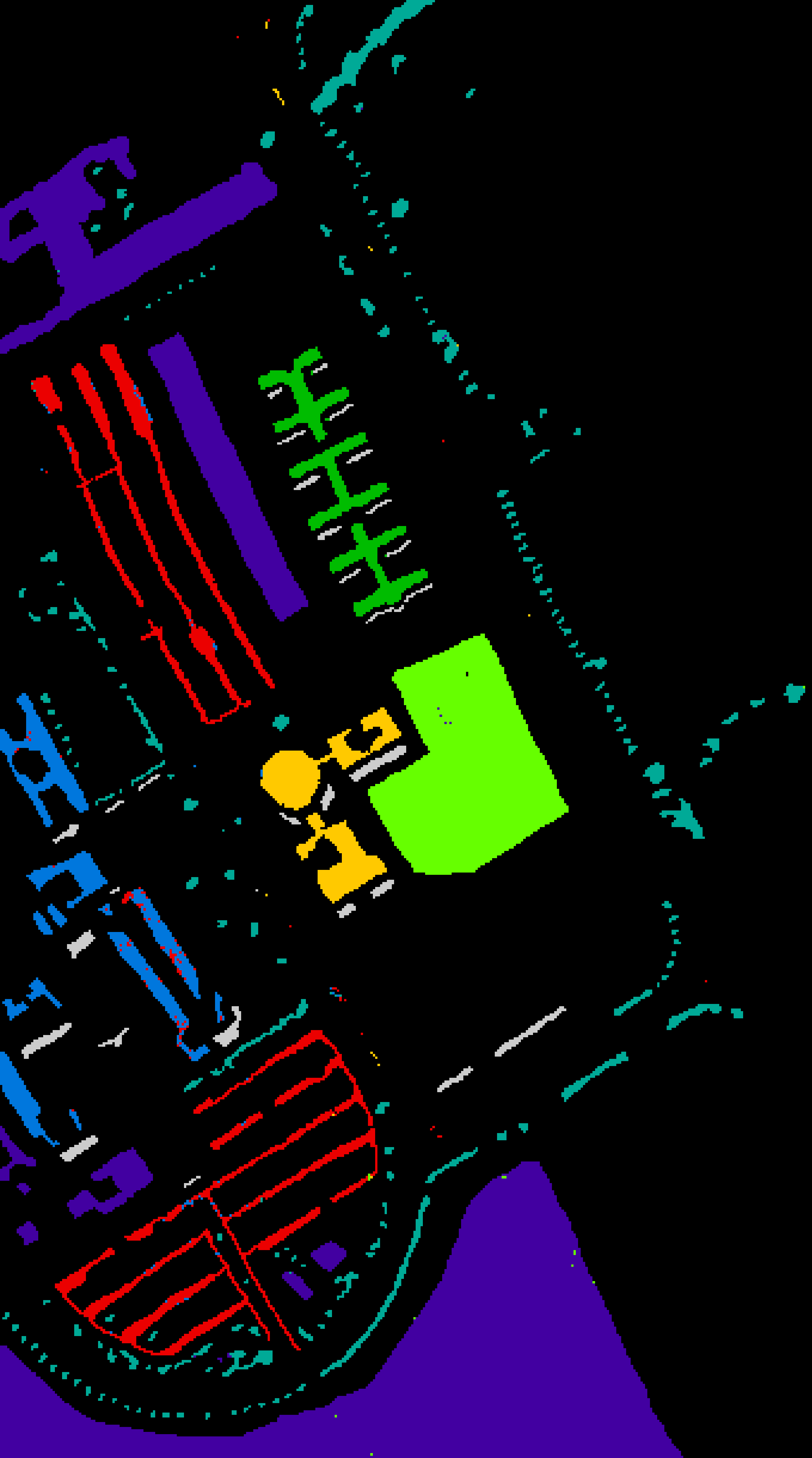}
	\caption*{Former}
    \end{subfigure}
    \begin{subfigure}{0.11\textwidth}
	\includegraphics[width=0.99\textwidth]{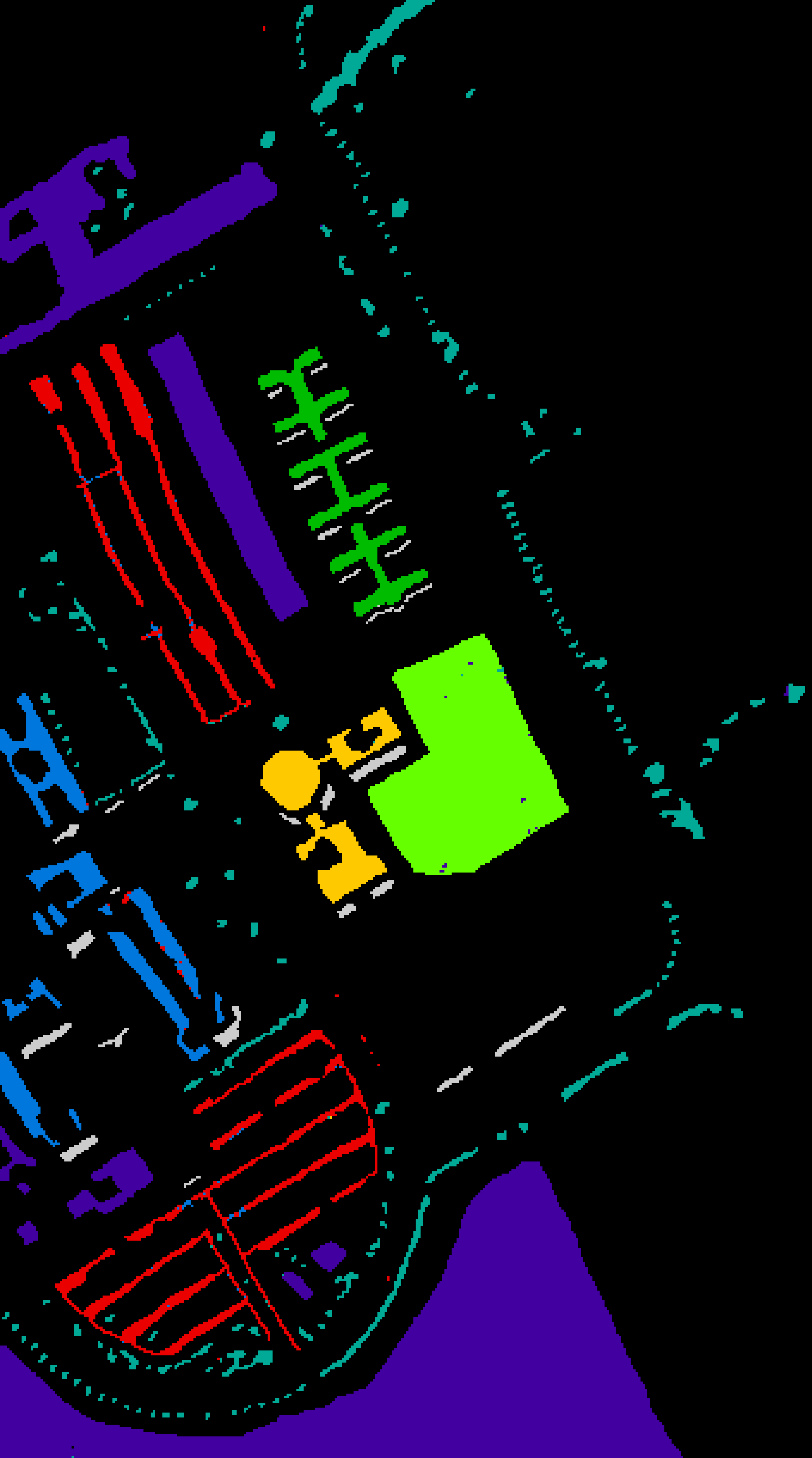}
	\caption*{PyFormer}
    \end{subfigure} 
    \begin{subfigure}{0.11\textwidth}
	\includegraphics[width=0.99\textwidth]{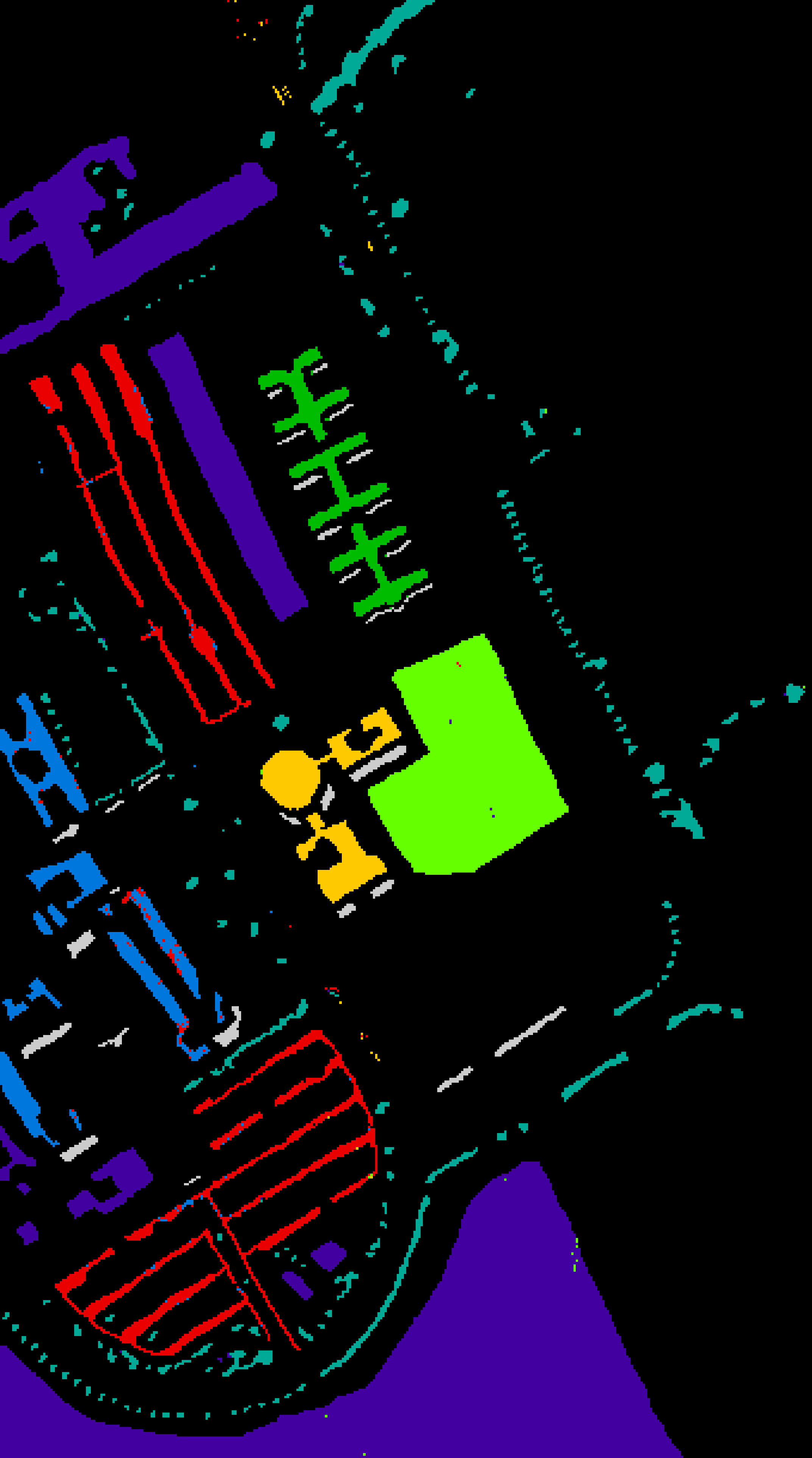}
	\caption*{WaveFormer}
    \end{subfigure}
    \begin{subfigure}{0.11\textwidth}
	\includegraphics[width=0.99\textwidth]{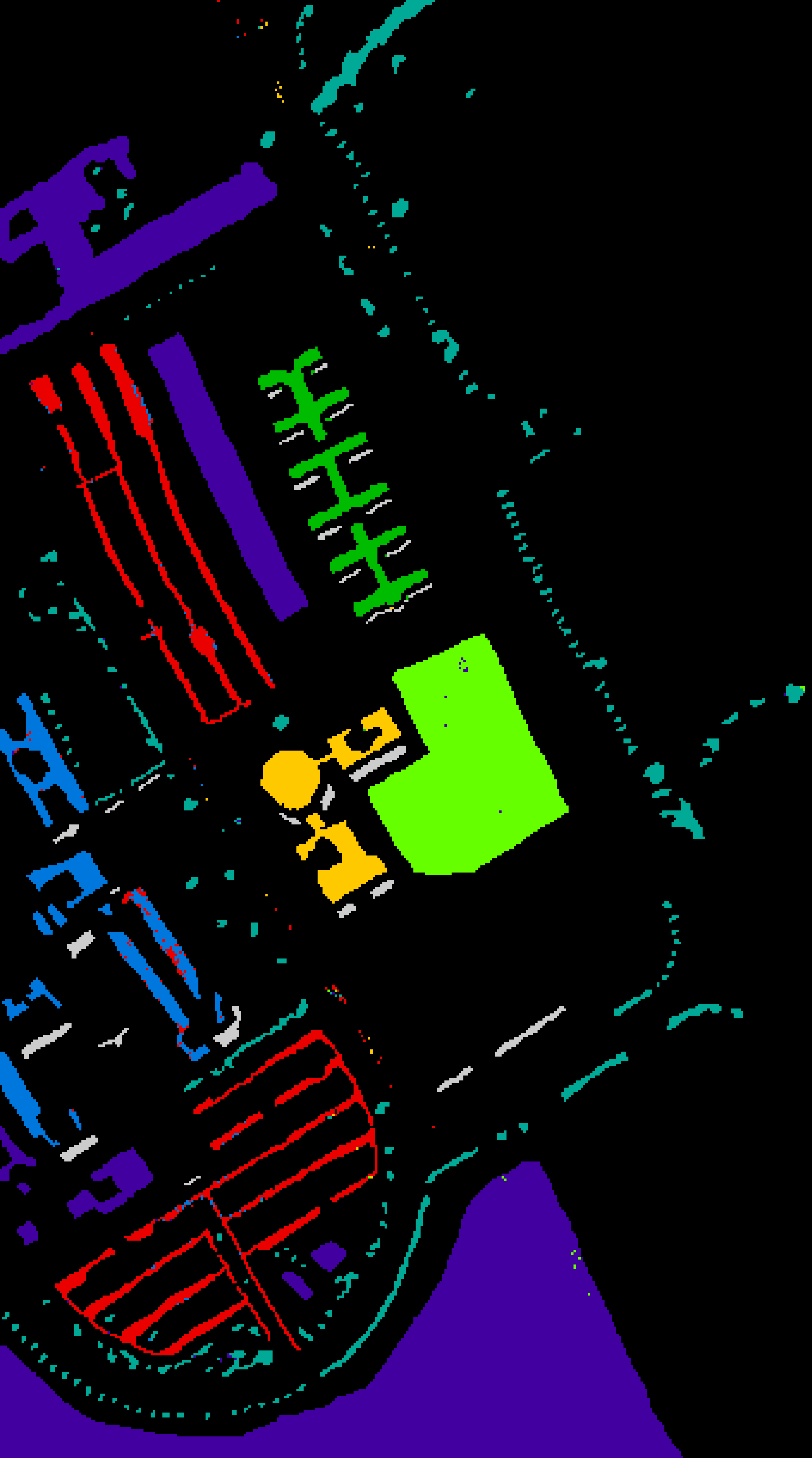}
	\caption*{HViT}
    \end{subfigure}
    \begin{subfigure}{0.11\textwidth}
	\includegraphics[width=0.99\textwidth]{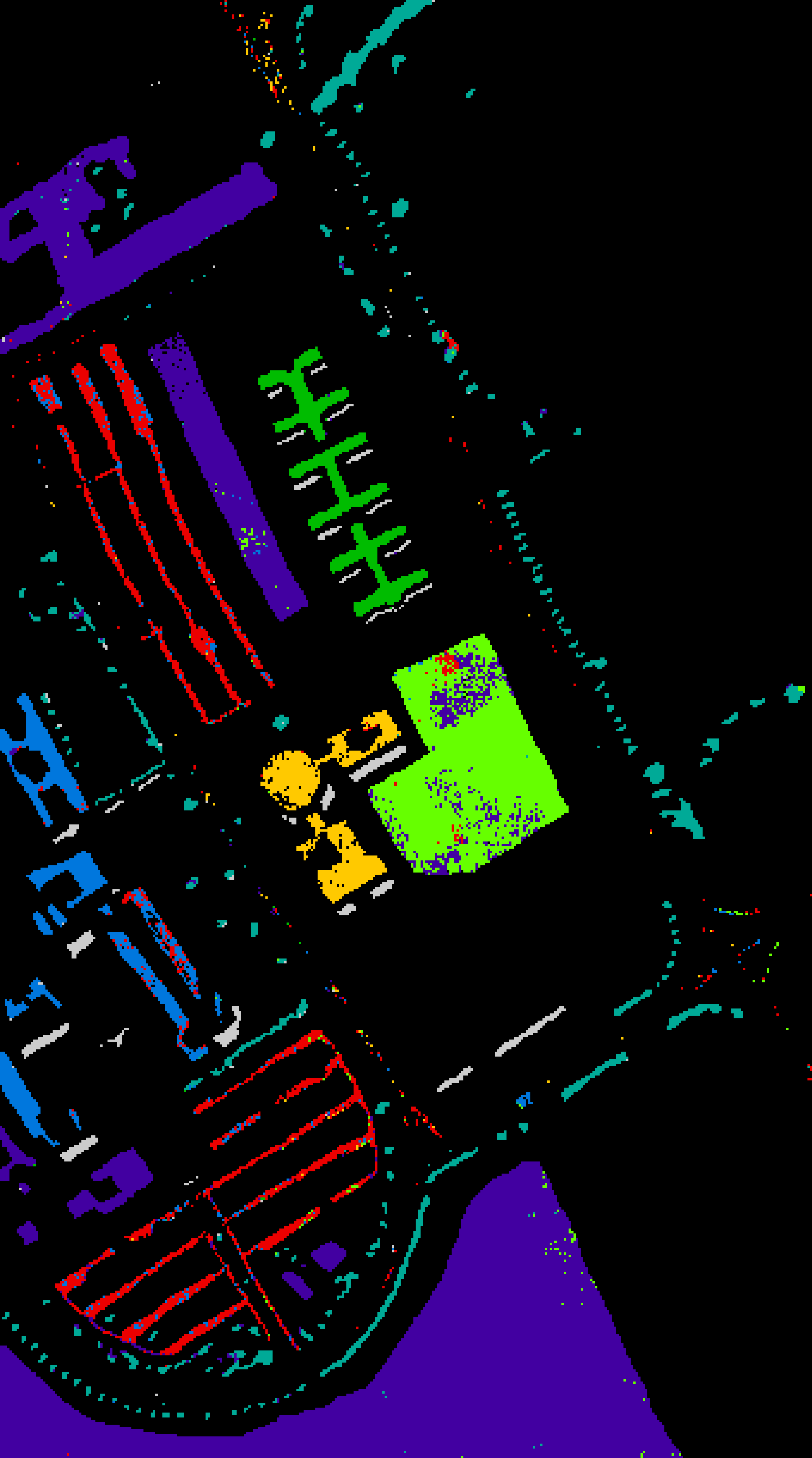}
	\caption*{MHMamba}
    \end{subfigure}
    \begin{subfigure}{0.11\textwidth}
	\includegraphics[width=0.99\textwidth]{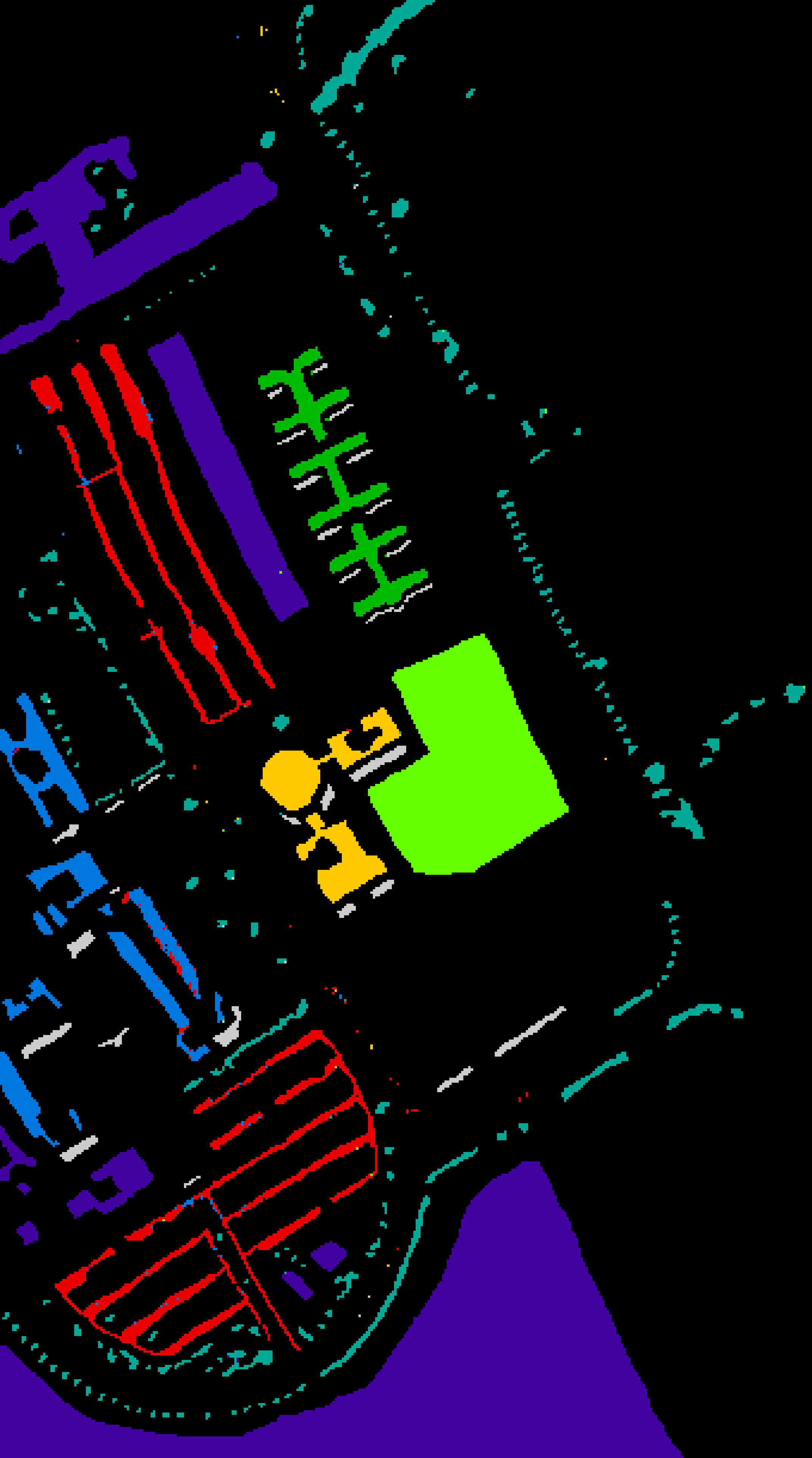}
	\caption*{WaveMamba}
    \end{subfigure}
    \begin{subfigure}{0.11\textwidth}
	\includegraphics[width=0.99\textwidth]{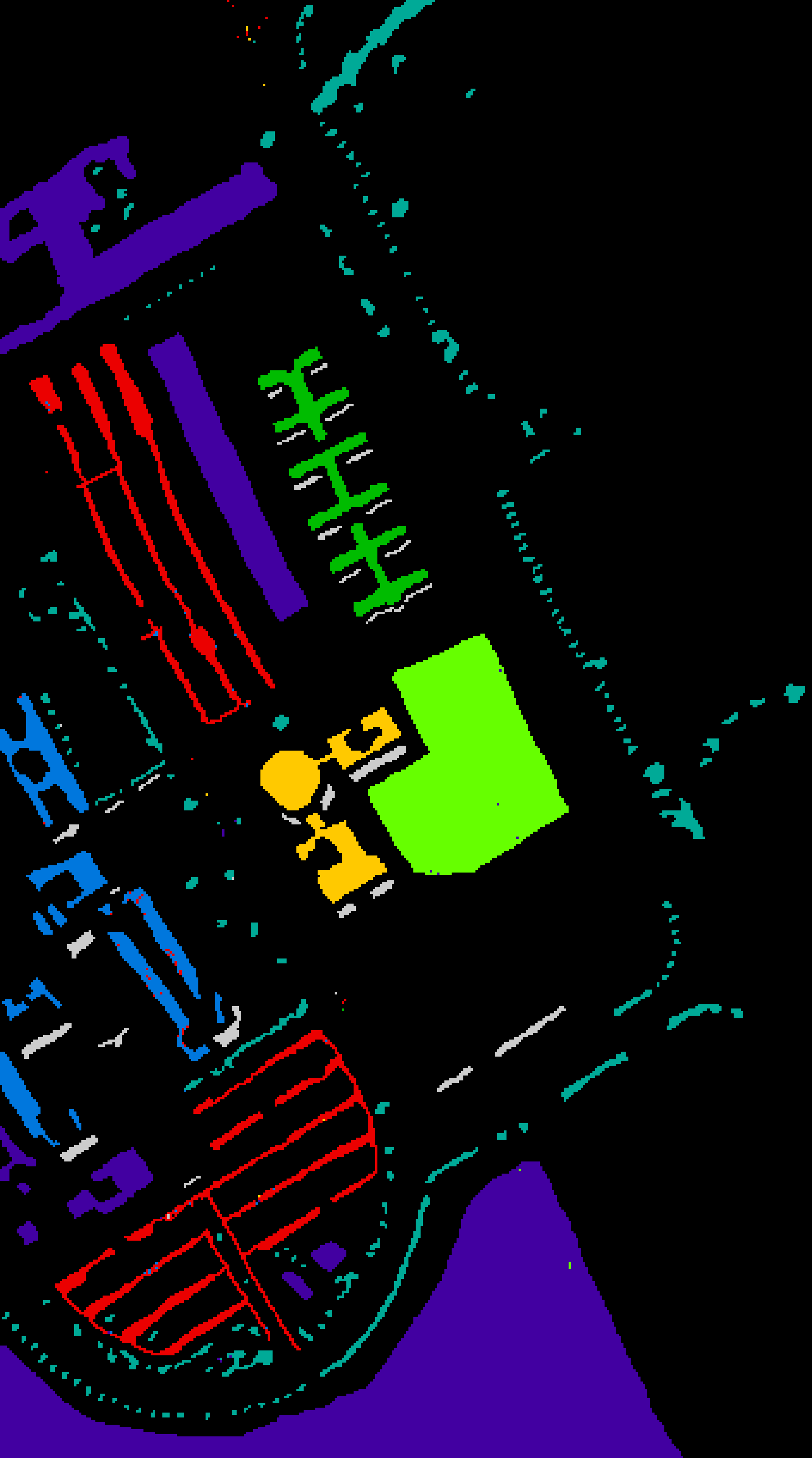}
	\caption*{DiffFormer}
    \end{subfigure}
\caption{Classification maps for the \textbf{PU dataset}, highlighting spatial variability and class-specific performance.}
\label{Fig10}
\end{figure}

Table \ref{Tab5} presents a comprehensive comparison of classification performance on the PU dataset across multiple models, using class-wise accuracies, aggregate metrics, and computational time as evaluation criteria. Notably, the proposed \textit{DiffFormer} demonstrates superior OA of 99.4623\%, $\kappa$ of 99.2872\%, and AA of 99.1498\%, consistently outperforming existing approaches. In particular, \textit{DiffFormer} achieves state-of-the-art results for challenging classes such as Asphalt (99.4972\%), Bitumen (99.7493\%), and Shadows (99.6478\%). These improvements suggest its effectiveness in capturing spatial and spectral correlations. In contrast, AGCN also achieves high performance, with the highest AA of 99.5006\%, but marginally lower OA and $\kappa$. The Former, PyFormer, and WaveFormer exhibit competitive results across most metrics but fall behind the proposed \textit{DiffFormer}, especially in the Bare Soil and Bitumen classes. WaveMamba and MHMamba, while innovative, exhibit suboptimal performance, particularly in classes such as Bricks and Gravel, indicating potential limitations in capturing finer spatial features.

From a computational perspective, \textit{DiffFormer} balances accuracy with efficiency, achieving competitive inference time (601.86s) relative to Former (383.60s) and WaveFormer (383.09s), while significantly outperforming models like PyFormer (2840.00s) and WaveMamba (3952.37s). This balance highlights its practical applicability for real-world scenarios requiring high accuracy and moderate computational overhead.

The classification maps in Figure \ref{Fig10} visually highlight the spatial distribution of predicted labels for each model. \textit{DiffFormer} showcases superior delineation of class boundaries and reduction in misclassified pixels, especially in heterogeneous regions such as Meadows and Bare Soil. In comparison, AGCN and WaveFormer exhibit satisfactory classification but suffer from subtle boundary inconsistencies. Models such as MHMamba and WaveMamba display noticeable artifacts and reduced precision in high-variability regions, consistent with their lower quantitative performance. Conversely, Former and PyFormer demonstrate balanced accuracy across large homogeneous classes but struggle in preserving spatial details in smaller regions like Bitumen and Bricks. The visualization further validates the effectiveness of \textit{DiffFormer} in handling spectral-spatial variability, ensuring high classification accuracy across diverse land cover types. The alignment between the quantitative metrics in Table \ref{Tab5} and the qualitative outputs in Figure \ref{Fig10} underscores the robustness and generalizability of the proposed approach.

\section{Conclusions and Future Research Directions}

This paper proposed \textit{DiffFormer}, a Differential Spatial-Spectral Transformer framework designed to tackle the challenges of HSIC, such as spectral redundancy and spatial discontinuity. The key innovation of this framework lies in its Differential Multi-Head Self-Attention (DMHSA) mechanism, which enhances hyperspectral feature representation by capturing subtle variations in spectral-spatial patches. By integrating SWiGLU activation and class token-based aggregation, \textit{DiffFormer} achieves efficient and accurate representation learning while maintaining computational efficiency. The patch-based spectral-spatial tokenization strategy further reduces input dimensionality without compromising essential information, making it scalable for large-scale datasets. Extensive experiments on benchmark hyperspectral datasets demonstrate that \textit{DiffFormer} outperforms SOTA models in classification accuracy and computational efficiency. While \textit{DiffFormer} demonstrates significant advancements in HSIC, several avenues for future research remain open for instance Investigate dynamic or hierarchical tokenization strategies to enable fine-grained spatial-spectral feature extraction without increasing computational overhead and Develop energy-efficient transformer architectures tailored for edge devices to extend the practical applications of \textit{DiffFormer} to UAVs and portable sensing systems.
\bibliographystyle{IEEEtran}
\bibliography{IEEEabrv,Sam}
\end{document}